# Stochastic Primal-Dual Methods and Sample Complexity of Reinforcement Learning

**Yichen Chen**  YICHENC@PRINCETON.EDU
*Department of Computer Science*
*Princeton University*
*Princeton, NJ 08544, USA*

**Mengdi Wang**  MENGDIW@PRINCETON.EDU
*Department of Operation Research and Financial Engineering*
*Princeton University*
*Princeton, NJ 08544, USA*

**Editor:**

## Abstract

We study the online estimation of the optimal policy of a Markov decision process (MDP). We propose a class of Stochastic Primal-Dual (SPD) methods which exploit the inherent minimax duality of Bellman equations. The SPD methods update a few coordinates of the value and policy estimates as a new state transition is observed. These methods use small storage and has low computational complexity per iteration. The SPD methods find an absolute-$\epsilon$-optimal policy, with high probability, using $\mathcal{O}\left(\frac{|\mathcal{S}|^4|\mathcal{A}|^2\sigma^2}{(1-\gamma)^6\epsilon^2}\right)$ iterations/samples for the infinite-horizon discounted-reward MDP and $\mathcal{O}\left(\frac{|\mathcal{S}|^4|\mathcal{A}|^2 H^6\sigma^2}{\epsilon^2}\right)$ for the finite-horizon MDP.

## 1. Introduction

Markov decision process (MDP) is one of the most basic model of dynamic programming, stochastic control and reinforcement learning; see the textbook references (Bertsekas, 1995; Bertsekas and Tsitsiklis, 1995; Sutton and Barto, 1998; Puterman, 2014). Given a controllable Markov chain and the distribution of state-to-state transition reward, the aim is to find the optimal action to perform at each state in order to maximize the expected overall reward. MDP and its numerous variants are widely applied in engineering systems, artificial intelligence, e-commerce and finance. Classical solvers of MDP require full knowledge of the underlying stochastic process and reward distribution, which are often not available in practice.

In this paper, we study both the infinite-horizon discounted-reward MDP and the finite-horizon MDP. In both cases, we assume that the MDP has a finite state space $\mathcal{S}$ and a finite action space $\mathcal{A}$. We focus on the model-free learning setting where both transition probabilities and transitional rewards are unknown. Instead, a simulation oracle is available to generate random state-to-state transitions and transitional rewards. The simulation oracle is able to model offline retrieval of static empirical data as well as live interaction with real-time simulation systems. The algorithmic goal is to estimate the optimal policy of the unknown MDP based on empirical state transitions, without any prior knowledge or restrictive assumption about the underlying process. In the literature of approximate dynamic programming and reinforcement learning, many methods have been developed and some of them are proved to achieve near-optimal performance guarantees in certain senses; recent examples including (Ortner and Auer, 2007; Dann and Brunskill, 2015; Lattimore et al., 2013; Lattimore and Hutter, 2012; Chen and Wang, 2016). Although researchers have made significant progress in developing reinforcement learning methods, it remains unclear whether there is an approach that achieves both theoretical optimality as well as practical scalability. This is an active area of research.

In this paper, we present a novel approach motivated by the linear programming formulation of the nonlinear Bellman equation. We formulate the Bellman equation into a stochastic saddle point problem, where the optimal primal and dual solutions correspond to the optimal value and policy functions, respectively. We propose a class of Stochastic Primal-Dual algorithms (SPD) for the discounted MDP and the finite-horizon MDP. Each iteration of the algorithms updates the primal and dual solutions simultaneously using noisy partial derivatives of the Lagrangian function. We show that one can compute a noisy partial derivative easily from a single observation of the state transition. The SPD methods are stochastic analogs of the primal-dual iteration for linear programming. They also involve alternative projections onto specially constructed sets. The SPD methods are very easy to implement and exhibit favorable space complexity. To analyze its sample complexity, we adopt the notion of "Probably Approximately Correct" (PAC), which means to achieve an $\epsilon$-optimal policy with high probability using sample size polynomial with respect to the problem parameters.

The main contributions of this paper are four-folded:

1. We study the basic linear algebra of reinforcement learning. We show that the optimal value and optimal policy are dual to each other, and they are the solutions to a stochastic saddle point problem. The *value-policy duality* implies convenient algebraic structure that may facilitate efficient learning and dimension reduction.

2. We develop a class of stochastic primal-dual (SPD) methods which maintain a value estimate and a policy estimate and update their coordinates while processing state-transition data incrementally. The SPD methods exhibit superior space and computational scalability. They require $\mathcal{O}(|\mathcal{S}| \times |\mathcal{A}|)$ space for discounted MDP and $\mathcal{O}(|\mathcal{S}| \times |\mathcal{A}| \times H)$ space for finite-horizon MDP. The space complexity of SPD is sublinear with respect to the input size of the MDP model. For discounted MDP, each iteration updates two coordinates of the value estimate and a single coordinate of the policy estimate. For finite-horizon MDP, each iteration updates $2H$ coordinates of the value estimate and $H$ coordinates of the policy estimate.

3. For discounted MDP, we develop the SPD-dMDP Algorithm 1. It yields an $\epsilon$-optimal policy with probability at least $1-\delta$ using the following sample size/iteration number

$$\mathcal{O}\left(\frac{|\mathcal{S}|^4 |\mathcal{A}|^2 \sigma^2}{(1-\gamma)^6 \epsilon^2} \ln\left(\frac{1}{\delta}\right)\right),$$

where $\gamma \in (0,1)$ is the discount factor, $|\mathcal{S}|$ and $|\mathcal{A}|$ are the sizes of the state space and the action space, $\sigma$ is a uniform upperbound of state-transition rewards. We obtain the sample complexity results by analyzing the duality gap sequence and applying the Bernstein inequality to a specially constructed martingale. The analysis is novel to the authors' best knowledge.

4. For finite-horizon MDP, we develop the SPD-fMDP Algorithm 2. It yields an $\epsilon$-optimal policy with probability at least $1 - \delta$ using the following sample size/iteration number

$$\mathcal{O}\left(\frac{|\mathcal{S}|^4 |\mathcal{A}|^2 H^6 \sigma^2}{\epsilon^2} \ln\left(\frac{1}{\delta}\right)\right),$$

where $H$ is the total number of periods. The key aspect of the finite-horizon algorithm is to adapt the learning rate/stepsize for updates on different periods. In particular, the algorithm has to update the policies associated with the earlier periods more aggressively than update those associated with the later periods.

The SPD is a model-free method and applies to a wide class of dynamic programming problems. Within the scope of this paper, the sample transitions are drawn from a static distribution. We conjecture that the sample complexity results can be improved by allowing exploitation, i.e., adaptive sampling of actions. The results of this paper suggest that the linear duality of MDP bears convenient structures yet to be fully exploited.

**Paper Organization** Section 2 reviews the basics of discounted and finite-horizon MDP and related works in this area. Section 3 studies the duality between optimal values and policies. Section 4 presents the SPD-dMDP and the SPD-fMDP algorithms and discuss their implementation and complexities. Section 5 presents the main results and Section 6 gives the proofs.

**Notations** All vectors are considered as column vectors. For a vector $x \in \Re^n$, we denote by $x^T$ its transpose, and denote by $\|x\| = \sqrt{x^T x}$ its Euclidean norm. For a matrix $A \in \Re^{n \times n}$, we denote by $\|A\| = \max\{\|Ax\| \mid \|x\| = 1\}$ its induced Euclidean norm. For a set $\mathcal{X} \subset \Re^n$ and vector $y \in \Re^n$, we denote by $\Pi_{\mathcal{X}}\{y\} = \operatorname{argmin}_{x \in \mathcal{X}} \|y - x\|^2$ the Euclidean projection of $y$ on $\mathcal{X}$, where the minimization is always uniquely attained if $\mathcal{X}$ is nonempty, convex and closed. We denote by $e = (1, \ldots, 1)^T$ the vector with all entries equaling 1, and we denote by $e_i = (0, ..., 0, 1, 0, ..., 0)^T$ the vector with its $i$-th entry equaling 1 and other entries equaling 0. For set $\mathcal{X}$, we denote its cardinality by $|\mathcal{X}|$.

## 2. Background

In this section, we review the basic models of Markov decision processes. We also survey existing literatures that are related to this paper.

### 2.1 Discounted Markov Decision Process

We consider a discounted MDP described by a tuple $\mathcal{M} = (\mathcal{S}, \mathcal{A}, \mathcal{P}, r, \gamma)$, where $\mathcal{S}$ is a finite state space, $\mathcal{A}$ is a finite action space, $\gamma \in (0,1)$ is a discount factor. If action $a$ is selected while the system is in state $i$, the system transitions to state $j$ with probability $P_a(i,j)$ and incurs a random reward $\hat{r}_{ija} \in [0, \sigma]$ with expectation $r_{ija}$.

Let $\pi : \mathcal{S} \mapsto \mathcal{A}$ be a *policy* that maps a state $i \in \mathcal{S}$ to an action $\pi(i) \in \mathcal{A}$. Consider the Markov chain under policy $\pi$. We denote its transition probability matrix by $P_\pi$ and denote its transitional reward vector by $r_{\pi^*}$, i.e.,

$$P_\pi(i,j) = P_{\pi(i)}a(i,j), \qquad r_\pi(i) = \sum_{j \in \mathcal{S}} P_{\pi(i)}(i,j) r_{ij\pi(i)}, \qquad i,j \in \mathcal{S}.$$

The objective is to find an optimal policy $\pi^* : \mathcal{S} \mapsto \mathcal{A}$ such that the infinite-horizon discounted reward is maximized, regardless of the initial state:

$$\max_{\pi:\mathcal{S} \mapsto \mathcal{A}} \mathbf{E}^\pi \left[ \sum_{k=0}^\infty \gamma^k \hat{r}_{i_k i_{k+1} \pi(i_k)} \right],$$

where $\gamma \in (0,1)$ is a discount factor, $(i_0, i_1, \ldots)$ are state transitions generated by the Markov chain under policy $\pi$, and the expectation is taken over the entire process. We assume throughout that *there exists a unique optimal policy $\pi^*$ to the MDP tuple $\mathcal{M} = (\mathcal{S}, \mathcal{A}, P, r, \gamma)$*. In other words, there exists one optimal action for each state.

We review the standard definitions of value functions.

**Definition 1** *The value vector $v^\pi \in \Re^{|\mathcal{S}|}$ of a fixed policy $\pi$ is defined as*

$$v^\pi(i) = \mathbf{E}^\pi \left[ \sum_{k=0}^\infty \gamma^k \hat{r}_{i_k i_{k+1} \pi(i_k)} \,\Big|\, i_0 = i \right], \qquad i \in \mathcal{S}.$$

**Definition 2** *The optimal value vector $v^* \in \Re^{|\mathcal{S}|}$ is defined as*

$$v^*(i) = \max_{\pi:\mathcal{S} \mapsto \mathcal{A}} \mathbf{E}^\pi \left[ \sum_{k=0}^\infty \gamma^k \hat{r}_{i_k i_{k+1} \pi(i_k)} \,\Big|\, i_0 = i \right] = \mathbf{E}^{\pi^*} \left[ \sum_{k=0}^\infty \gamma^k \hat{r}_{i_k i_{k+1} \pi^*(i_k)} \,\Big|\, i_0 = i \right], \qquad i \in \mathcal{S}.$$

For the sample complexity analysis of the proposed algorithm, we need a notion of sub-optimality of policies. We give its definition as below.

**Definition 3** *We say that a policy $\pi$ is absolute-$\epsilon$-optimal if*

$$\max_{i \in \mathcal{S}} |v^\pi(i) - v^*(i)| \leq \epsilon.$$

If a policy is absolute-$\epsilon$-optimal, it achieves $\epsilon$-optimal reward regardless of the initial state distribution. We note that the absolute-$\epsilon$-optimality is one of the strongest notions of sub-optimality for policies. In comparison, some literatures analyze the expected sub-optimality of a policy when the state $i$ follows a certain distribution.

## 2.2 Finite-Horizon Markov Decision Process

We also consider the finite-horizon Markov decision process, which can be formulated as a tuple $\mathcal{M} = (\mathcal{S}, \mathcal{A}, H, \mathcal{P}, r)$, where $\mathcal{S}$ is a finite state space with transition probabilities encoded by $\mathcal{P} = (P_a)_{a \in \mathcal{A}} \in \mathcal{R}^{|\mathcal{S} \times \mathcal{S} \times \mathcal{A}|}$, $\mathcal{A}$ is a finite action space and $H$ is the horizon. If action $a$ is selected while the system is in state $i \in \mathcal{S}$ at period $h = 0, \ldots, H-1$, the system transitions to state $j$ and period $h+1$ with probability $P_a(i, j)$ and incurs a random reward $\hat{r}_{ija}$ with expectation $r_{ija}$. For the simplicity of presentation, we assume that the reward we receive does not depend on the time period. Our algorithm can be readily extended to the case when the reward varies with the time period. We assume that both $\mathcal{P}$ and $r$ are unknown but they can be estimated by sampling.

We augment the state space with the time period to obtain an augmented Markov chain. Now we have a replica of $\mathcal{S}$ at each period $h$, denoted by $\mathcal{S}_h$. Let $\mathcal{S}_{[H]}$ be the state space of the augmented MDP where $[H]$ denotes the set of integers $\{0, 1, \ldots, H-1\}$. If we select action $a$ in state $(i, h) \in \mathcal{S}_{[H]}$, the state transitions to a new state $(j, h+1) \in \mathcal{S}_{[H]}$ with probability $P_a(i, j)$. The transition incurs a random reward $\hat{r}_{ija} \in [0, \sigma]$ with expectation $r_{ija}$. At period $H-1$, the state $i$ will transition to the terminal state with reward $\hat{r}_{ija}$. In the rest of the paper, we use $\Pi_a(i', j')$ to denote the transition probability of the augmented Markov chain where $i', j' \in \{(i, h) | i \in \mathcal{S}, h \in [H]\}$.

Let $\pi = (\pi_0, \ldots, \pi_{H-1})$ be a sequence of *one-step policies* such that $\pi_h$ maps a state $i \in \mathcal{S}$ to an action $\pi_h(i) \in \mathcal{A}$ in the $h$-th period. Consider the augmented Markov chain under policy $\pi$. We denote its transition probability matrix by $\Pi_\pi$, where $\Pi_\pi((i,h), (j, h+1)) = P_{\pi_h(i)}(i, j)$ for all $h \in [H-1]$, $i, j \in \mathcal{S}$, which is given by

$$\Pi_\pi = \begin{bmatrix} 0 & P_{\pi_0} & 0 & \ldots & 0 \\ 0 & 0 & P_{\pi_1} & \ldots & 0 \\ \vdots & \vdots & \vdots & \ddots & \vdots \\ 0 & 0 & 0 & \ldots & P_{\pi_{H-2}} \\ 0 & 0 & 0 & \ldots & 0 \end{bmatrix}.$$

We denote $r_a \in \Re^{|\mathcal{S}|}$ to be the expected state transition reward under action $a$ such that

$$r_a(i) = \sum_{j \in \mathcal{S}} P_a(i, j) r_{ija}, \quad i \in \mathcal{S}.$$

We review the standard definitions of value functions for finite-horizon MDP.

**Definition 4** *The $h$-period value function $v_h^\pi : \mathcal{S}_h \mapsto \Re^n$ under policy $\pi$ is defined as*

$$v_h^\pi(i) = \mathbf{E}^\pi \left[ \sum_{\tau=h}^{H-1} \hat{r}_{i_\tau i_{\tau+1} \pi_\tau(i_\tau)} \middle| i_h = i \right], \forall\ h \in [H], i \in \mathcal{S}.$$

The random variables $(i_h, i_{h+1}, \ldots)$ are state transitions generated by the Markov chain under policy $\pi$ starting from state $i$ and period $h$, and the expectation $\mathbf{E}^\pi[\cdot]$ is taken over the entire process. We denote the overall value function to be $v = (v_0^T, \ldots, v_{H-1}^T)^T \in \mathbb{R}^{|\mathcal{S}|H}$.

The objective of the finite-horizon MDP is to find an optimal policy $\pi^* : \mathcal{S}_{[H]} \mapsto \mathcal{A}$ such that the finite-horizon reward is maximized, regardless of the starting state. Based on the optimal policy, we define the *optimal value function* as follows.

**Definition 5** *The optimal value vector $v^* = (v_h^*)_{h \in [H]} \in \Re^{H \times |\mathcal{S}|}$ is defined as*

$$v_h^*(i) = \max_{\pi: \mathcal{S}_{[H]} \mapsto \mathcal{A}} \mathbf{E}^\pi \left[ \sum_{\tau=h}^{H-1} \hat{r}_{i_\tau i_{\tau+1} \pi_\tau(i_\tau)} \bigg| i_h = i \right] = \mathbf{E}^{\pi^*} \left[ \sum_{\tau=h}^{H-1} \hat{r}_{i_\tau i_{\tau+1} \pi_\tau(i_\tau)} \bigg| i_h = i \right],$$

*for all $h \in [H], i \in \mathcal{S}$.*

In order to analyze the sample complexity, we use the following notion of *absolute-$\epsilon$-optimal* for finite-horizon MDP.

**Definition 6** *We say that a policy $\pi$ is absolute-$\epsilon$-optimal if*

$$\max_{h \in [H], i \in \mathcal{S}} |v_h^\pi(i) - v_h^*(i)| \leq \epsilon.$$

Note that if a policy is absolute-$\epsilon$-optimal, it achieves an $\epsilon$-optimal cumulative reward from all states and all intermediate periods. This is one of the strongest notions of suboptimality for finite-horizon policies.

### 2.3 Related Works

Our proposed methods use ideas from the linear program approach for Bellman's equations and the stochastic approximation method. The linear program formulation of Bellman's equation was known at around the same time when Bellman's equation was known; see Bertsekas (1995); Puterman (2014). Ye (2011) shows that the policy iteration of discounted MDP is a form of dual simplex method, which is strongly polynomial for the equivalent linear program and terminates in run time $\mathcal{O}(\frac{|\mathcal{A} \times \mathcal{S}|^2}{1-\gamma})$. Cogill (2015) considered the exact primal-dual method for MDP with full knowledge and interpreted it as a form of value iteration. Approximate linear programming approaches have been developed for solving the large-scale MDP on a low-dimensional subspace, started with de Farias and Van Roy (2003) and followed by Veatch (2013); Abbasi-Yadkori et al. (2014).

Our algorithm and analysis is closely related to the class of stochastic approximation (SA) methods. For textbook references on stochastic approximation, please see (Kushner and Yin, 2003; Benveniste et al., 2012; Borkar, 2008; Bertsekas and Tsitsiklis, 1989). We also use the averaging idea by Polyak and Juditsky (1992). In particular, our algorithm can be viewed as a stochastic approximation method for stochastic saddle point problems, which was first studied by Nemirovski and Rubinstein (2005) without rate of convergence analysis. Recent works (Dang and Lan, 2014; Chen et al., 2014) studied stochastic first-order methods for a class of general stochastic convex-concave saddle point problems and obtained optimal and near-optimal convergence rates.

In the literatures of reinforcement learning, there have been works on dual temporal difference learning which use primal-dual-type methods, see for examples (Yang et al., 2009; Liu et al., 2012; Mahadevan and Liu, 2012; Mahadevan et al., 2014). These works focused on evaluating the value function for a *fixed policy*. This is different from our work where the aim is to learn the *optimal policy*. We also remark that a primal-dual learning method has been considered for the optimal stopping problem by Borkar et al. (2009). However, no explicit sample complexity analysis is available.

In the recent several years, a growing body of work has provided various reinforcement learning methods that are able to "learn" the optimal policy with sample complexity guarantees. The notion of "Probably Approximately Correct" (PAC) was considered for MDP by Strehl et al. (2009), which requires that the learning method output an $\epsilon$-optimal policy using sample size that is polynomial with respect to parameters of the MDP and $1/\epsilon$ with high probability. Since then, many methods have been developed for discounted MDP and proved to achieve various PAC guarantees. Strehl et al. (2009) showed that the R-MAX has sample complexity $\mathcal{O}(S^2A/(\epsilon^3(1-\gamma)^6))$ and Delayed Q-learning has $\mathcal{O}(SA/(\epsilon^4(1-\gamma)^8))$. Lattimore and Hutter (2012) proposed the Upper Confidence Reinforcement Learning and obtained a PAC upper and lower bound $\mathcal{O}\left(\frac{SA}{\epsilon^2(1-\gamma)^3}\log\frac{1}{\delta}\right)$ under a restrictive assumption: one can only move to two states from any given state. Lattimore et al. (2013) extended the analysis to more general RL models. Azar et al. (2012) showed that the model-based value iteration achieves the optimal rate $\mathcal{O}\left(\frac{|\mathcal{S}\times\mathcal{A}|\log(|\mathcal{S}\times\mathcal{A}|/\delta)}{(1-\gamma)^3\epsilon^2}\right)$ for discounted MDP. Dann and Brunskill (2015) developed a upper confidence method for fixed-horizon MDP and obtain a near-optimal rate $\mathcal{O}\left(\frac{S^2AH^2}{\epsilon^2}\ln\frac{1}{\delta}\right)$. They also provide a lower bound $\Omega(\frac{SAH^2}{\epsilon^2}\ln\frac{1}{\delta+c})$. Based on their PAC bounds, Osband and Van Roy (2016) conjecture that the regret lower bound for reinforcement learning is $\Omega(\sqrt{SAT})$. Although the upper confidence methods achieve the close-to-optimal PAC complexity in some cases, they require maintaining a confidence interval for each state-action-state triplet. Thus these methods are not yet satisfactory in the space complexity. It remains unclear whether there is a approach that achieves both the space efficiency and the near-optimal sample complexity guarantee, without estimating the transition probabilities. This motivates the research in this paper.

We emphasize that the SPD methods proposed in this paper differ fundamentally from the existing methods mentioned above. In contrast, the SPD methods are more closely related to stochastic first-order methods for convex optimization and convex-concave saddle point problems. The closest prior work to the current one is (Chen and Wang, 2016), which proposed the first primal-dual-type learning method and a loose error bound. Later the work Chen et al. (2017) proposed a primal-dual method for the finite horizon MDP. No PAC analysis was given in (Chen and Wang, 2016; Chen et al., 2017). In this paper, we develop a new class of stochastic primal-dual methods, which are substantially improved in both practical efficiency and theoretical complexity. Practically, the new algorithms are essentially coordinate descent algorithms involving projections onto simple sets. As a result, each iteration is very easy to implement, making the algorithms practically favorable. Theoretically, these methods come with rigorous sample complexity guarantees. The results of this paper provide the first PAC guarantee for primal-dual reinforcement learning.

## 3. Value-Policy Duality of Markov Decision Processes

In this section, we study the Bellman equation of the Markov decision process from the perspective of linear duality. We show that the optimal value and policy functions are dual to each other and they are the solutions to a special saddle point problem. We analyze the value-policy duality for the infinite-horizon discounted-reward case and the finite-horizon case separately.

### 3.1 Case I: Infinite-Horizon Discounted-Reward Markov Decision Processes

Consider the discounted MDP described by the tuple $\mathcal{M} = (\mathcal{S}, \mathcal{A}, \mathcal{P}, r, \gamma)$ as in Section 2.1. According to the theory of dynamic programming (Bertsekas, 1995), a vector $v^*$ is the optimal value function to the MDP if and only if it satisfies the following $|\mathcal{S}| \times |\mathcal{S}|$ system of equations, known as the *Bellman equation*, given by

$$v^*(i) = \max_{a \in \mathcal{A}} \left\{ \gamma \sum_{j \in \mathcal{S}} P_a(i,j) v^*(j) + \sum_{j \in \mathcal{S}} P_a(i,j) r_{ija} \right\}, \quad i \in \mathcal{S}, \tag{1}$$

When $\gamma \in (0,1)$, the Bellman equation has a unique fixed point solution $v^*$, and it equals to the optimal value function of the MDP. Moreover, a policy $\pi^*$ is an optimal policy for the MDP if and only of it attains the minimization in the Bellman equation. Note that this is a *nonlinear* system of fixed point equations. Interestingly, the Bellman equation (1) is equivalent to the following $|\mathcal{S}| \times (|\mathcal{S}||\mathcal{A}|)$ linear programming (LP) problem (see Puterman (2014) Section 6.9. and the paper by de Farias and Van Roy (2003)):

$$\begin{aligned} &\text{minimize } \xi^T v \\ &\text{subject to } (\mathbf{I} - \gamma P_a) v - r_a \geq 0, \quad a \in \mathcal{A}, \end{aligned} \tag{2}$$

where $\xi$ is an arbitrary vector with positive entries, $P_a \in \Re^{|\mathcal{S}| \times |\mathcal{S}|}$ is matrix whose $(i,j)$-th entry equals to $P_a(i,j)$, $\mathbf{I}$ is the identity matrix with dimension $|\mathcal{S}| \times |\mathcal{S}|$ and $r_a \in \Re^{|\mathcal{S}|}$ is the expected state transition reward under action $a$ given by $r_a(i) = \sum_{j \in \mathcal{S}} P_a(i,j) r_{ija}, i \in \mathcal{S}$. The dual linear program of (2) is

$$\begin{aligned} &\text{maximize } \sum_{a \in \mathcal{A}} \lambda_a^T r_a \\ &\text{subject to } \sum_{a \in \mathcal{A}} \left( \mathbf{I} - \gamma P_a^T \right) \lambda_a = \xi, \quad \lambda_a \geq 0, a \in \mathcal{A}. \end{aligned} \tag{3}$$

We will show that the optimal solution $\lambda^*$ to the dual problem (3) corresponds to the optimal policy $\pi^*$ of the MDP. The duality between the optimal value vector and the optimal policy is established in Theorem 1. We remark that part of the results were known in the classical literature of MDP; see Puterman (2014) Section 6.9. We provide a short proof for the completeness of analysis.

**Theorem 1 (Value-Policy Duality for Discounted MDP)** *Assume that the discounted-reward infinite-horizon MDP tuple $\mathcal{M} = (\mathcal{S}, \mathcal{A}, P, r, \gamma)$ has a unique optimal policy $\pi^*$. Then $(v^*, \lambda^*)$ is the unique pair of primal and dual solutions to (2), (3) if and only if*

$$v^* = (\mathbf{I} - \gamma P_{\pi^*})^{-1} r_{\pi^*}, \qquad \left( \lambda^*_{\pi^*(i), i} \right)_{i \in \mathcal{S}} = (\mathbf{I} - \gamma (P_{\pi^*})^T)^{-1} \xi, \qquad \lambda^*_{a,i} = 0 \text{ if } a \neq \pi^*(i).$$

*Proof.* The proof is based on the fundamental property of linear programming, i.e., $(v^*, \lambda^*)$ is the optimal pair of primal and dual solutions if and only if:

(a) $v^*$ is primal feasible, i.e., $(\mathbf{I} - \gamma P_a) v^* - r_a \geq 0$ for all $a \in \mathcal{A}$.

(b) $\lambda^*$ is dual feasible, i.e., $\sum_{a \in \mathcal{A}} \left(\mathbf{I} - \gamma P_a^T\right) \lambda_a^* = \xi$ and $\lambda_a^* \geq 0$ for all $a \in \mathcal{A}$.

(c) $(v^*, \lambda^*)$ satisfies the complementarity slackness condition
$$\lambda_{a,i}^* \cdot (v_i^* - \gamma P_{a,i} v^* - r_{a,i}) = 0 \quad \forall i \in \mathcal{S}, a \in \mathcal{A},$$
where $\lambda_{a,i}^*$ is the $i$-th element of $\lambda_a^*$ and $P_{a,i}$ is the $i$-th row of $P_a$.

Suppose that $(v^*, \lambda^*)$ is primal-dual optimal. As a result, it satisfies (a), (b), (c) and $v^*$ is the optimal value vector. By the definition of the optimal value function, we know that $v_i^* - \gamma P_{\pi^*(i),i} v^* - r_{\pi^*(i),i} = 0$. Since $\pi^*$ is unique, we have $v_i^* > \gamma P_{a,i} v^* + r_{a,i} = 0$ if $a \neq \pi^*(i)$. As a result, we have $\lambda_{a,i}^* = 0$ for all $a \neq \pi^*(i)$. It means that the optimal dual variable $\lambda^*$ has exactly $|\mathcal{S}|$ nonzeros, corresponding to $|\mathcal{S}|$ active row constraints of the primal problem (2). We combine this observation with the dual feasibility relation $\sum_{a \in \mathcal{A}} \left(\mathbf{I} - \gamma P_a^T\right) \lambda_a^* = \xi$ and obtain
$$(\mathbf{I} - \gamma (P_{\pi^*})^T) \left(\lambda_{\pi^*(i),i}^*\right)_{i \in \mathcal{S}} = \xi.$$

Note that all eigenvalues of $P_{\pi^*}$ belong to the unit ball, therefore $(\mathbf{I} - \gamma (P_{\pi^*})^T)$ is invertible. Then we have $\left(\lambda_{\pi^*(i),i}^*\right)_{i \in \mathcal{S}} = (\mathbf{I} - \gamma (P_{\pi^*})^T)^{-1} \xi$, which together with the complementarity condition implies that $\lambda^*$ is unique. Similarly we can show that $v^* = (\mathbf{I} - \gamma P_{\pi^*})^{-1} r_{\pi^*}$ from the primal feasibility and slackness condition.

Now suppose that $(v^*, \lambda^*)$ satisfies the three conditions stated in Theorem 1. Then we obtain (a),(b),(c) directly, which proves that $(v^*, \lambda^*)$ is primal-dual optimal. ∎

Theorem 1 suggests a critical correspondence between the optimal dual solution $\lambda^*$ and the optimal policy $\pi^*$. In particular, one can recover the optimal policy $\pi^*$ from the basis of $\lambda^*$ as follows:
$$\pi^*(i) = a, \quad \text{if } \lambda_{a,i}^* > 0.$$
In other words, *finding the optimal policy is equivalent to finding the basis of the optimal dual solution.* This suggests that learning the optimal policy is a special case of stochastic linear programs.

**Saddle Point Formulation of Discounted MDP** We rewrite the LP program (2) into an equivalent minimax problem, given by
$$\min_{v \in \Re^{|\mathcal{S}|}} \max_{\lambda \geq 0} L(v, \lambda) = \xi^T v + \sum_{a \in \mathcal{A}} \lambda_a^T \left((\gamma P_a - \mathbf{I}) v + r_a\right). \quad (4)$$

The primal variable $v$ is of dimension $|\mathcal{S}|$, and the dual variable
$$\lambda = (\lambda_a)_{a \in \mathcal{A}} = (\lambda_{a,i})_{a \in \mathcal{A}, i \in \mathcal{S}}$$
is of dimension $|\mathcal{S}| \cdot |\mathcal{A}|$. Each subvector $\lambda_a \in \Re^{|\mathcal{S}|}$ is the vector multiplier corresponding to constraint inequalities $(\mathbf{I} - \gamma P_a) v - r_a \geq 0$. Each entry $\lambda_{a,i} > 0$ is the scalar multiplier associated with the $i$th row of $(\mathbf{I} - \gamma P_a) v - r_a \geq 0$.

In order to develop an efficient algorithm, we modify the saddle point problem as follows
$$\min_{v \in \Re^{|\mathcal{S}|}} \max_{\lambda \in \Re^{|\mathcal{S}| \times |\mathcal{A}|}} \left\{ L(v, \lambda) = \xi^T v + \sum_{a \in \mathcal{A}} \lambda_a^T \left((\gamma P_a - \mathbf{I}) v + r_a\right) \right\}, \quad (5)$$
$$\text{subject to} \quad v \in \mathcal{V}, \quad \lambda \in \Xi \cap \Delta$$

where
$$\mathcal{V} = \left\{ v \mid v \geq 0, \|v\|_\infty \leq \frac{\sigma}{1-\gamma} \right\},$$
$$\Xi = \left\{ \lambda \mid \sum_{a \in \mathcal{A}} \lambda_{a,i} \geq \xi_i, \ \forall i \in \mathcal{S} \right\}, \quad \Delta = \left\{ \lambda \mid \lambda \geq 0, \|\lambda\|_1 = \frac{\|\xi\|_1}{1-\gamma} \right\}. \tag{6}$$

We will show later that $v^*$ and $\lambda^*$ belong to $\mathcal{V}$ and $\Xi \cap \Delta$ respectively (Lemma 1). As a result, the modified saddle point problem (5) is equivalent to the original problem (4).

### 3.2 Case II: Finite-Horizon Markov Decision Processes

Consider the finite-horizon Markov decision process, which is described by a tuple $\mathcal{M} = (\mathcal{S}, \mathcal{A}, H, \mathcal{P}, r)$ as in Section 2.2. The Bellman equation of finite-horizon MDP is given by

$$v_h^*(i) = \max_{a \in \mathcal{A}} \left\{ r_a(i) + \sum_{j \in \mathcal{S}} P_a(i,j) v_{h+1}^*(j) \right\}, \forall\ h \in [H], i \in \mathcal{S}, \tag{7}$$
$$v_H^* = 0,$$

where $P_a$ is the transition probability matrix using a fixed action $a$. The vector form of the Bellman equation is

$$v_h^* = \max_{a \in \mathcal{A}} \left\{ r_a(i) + P_a v_{h+1}^* \right\}, \quad \forall\ h \in [H], i \in \mathcal{S},$$
$$v_H^* = 0,$$

where the maximization is carried out component-wise. A vector $v^*$ satisfies the Bellman equation if and only if it is the optimal value function.

The Bellman equation is equivalent to the following linear program:

$$\begin{aligned}
\text{minimize} \quad & \sum_{h=0}^{H-1} \xi_h^T v_h \\
\text{subject to} \quad & v_h \geq P_a v_{h+1} + r_a, \quad a \in \mathcal{A}, h \in [H] \\
& v_H = 0,
\end{aligned} \tag{8}$$

where $\xi = (\xi_0, \ldots, \xi_{H-1})$ is an aribrary vector with positive entries and $v = (v_0^T, \ldots, v_{H-1}^T)^T$ is the primal variable. The above linear program has $|\mathcal{S}||\mathcal{A}|H$ constraints and $|\mathcal{S}|H$ variables. The dual linear program of (8) is given by

$$\begin{aligned}
\text{maximize} \quad & \sum_{h=0}^{H-1} \sum_{a \in \mathcal{A}} \lambda_{h,a}^T r_a \\
\text{subject to} \quad & \sum_{a \in \mathcal{A}} \left( \lambda_{h,a} - P_a^T \lambda_{h-1,a} \right) = \xi_h, \quad h \in [H] \\
& \lambda_{-1,a} = 0, \ \lambda_{h,a} \geq 0, \quad h \in [H], a \in \mathcal{A}
\end{aligned} \tag{9}$$

where the dual variable $\lambda = (\lambda_{h,a})_{h \in [H], a \in \mathcal{A}}$ is a vector of dimension $|\mathcal{S}||\mathcal{A}|H$. In the remainder of the paper, we use the notation $\lambda_h$ to denote the vector $\lambda_h = (\lambda_{h,a})_{a \in \mathcal{A}}$ and $\lambda_a$

to denote the vector $\lambda_a = (\lambda_{h,a})_{h \in [H]}$. We denote by $\lambda^*$ the optimal dual solution. Now we establish the value-policy duality for finite-horizon MDP.

**Theorem 2 (Value-Policy Duality for Finite-Horizon MDP)** *Assume that the finite-horizon MDP tuple $\mathcal{M} = (\mathcal{S}, \mathcal{A}, H, \mathcal{P}, r)$ has a unique optimal policy $\pi^* = (\pi_h^*)_{h \in [H]}$. The vector pair $(v^*, \lambda^*)$ is the unique pair of primal and dual solutions to (8), (9) if and only if:*

$$v^* = (\mathbf{I} - \Pi_{\pi^*})^{-1}(R_{h,\pi^*})_{h \in [H]}, \quad \left(\lambda_{h,\pi_h^*(i)}^*(i)\right)_{h \in [H], i \in \mathcal{S}} = (\mathbf{I} - \Pi_{\pi^*}^T)^{-1}\xi, \quad \lambda_{h,a}^*(i) = 0 \text{ if } a \neq \pi_h^*(i),$$

*where $\Pi_{\pi^*}$ is the transition probability under the optimal policy $\pi^*$, $R_{h,\pi^*}$ is the expected transitional reward under policy, i.e., $R_{h,\pi^*}(i) = r_{\pi_h^*(i)}(i)$ for all $i \in \mathcal{S}$ and $h \in [H]$.*

*Proof.* The proof is based on the fundamental property of the linear duality, i.e., $(v^*, \lambda^*)$ is the optimal pair of primal and dual solutions if and only if

(a) $v^*$ is primal feasible, i.e., $v_h^* - P_a v_{h+1}^* - r_a \geq 0, \quad \forall h \in [H], a \in \mathcal{A}$.

(b) $\lambda^*$ is dual feasible, i.e., $\sum_{a \in \mathcal{A}} \left(\lambda_{h,a} - P_a^T \lambda_{h-1,a}\right) = \xi_h, \ \lambda_{h,a} \geq 0, \quad \forall h \in [H], a \in \mathcal{A}$.

(c) $(v^*, \lambda^*)$ satisfies the complementarity slackness condition

$$\lambda_{h,a}^*(i) \cdot (v_h^*(i) - r_a(i) - P_{a,i} v_{h+1}^*) = 0 \quad \forall h \in [H], i \in \mathcal{S}, a \in \mathcal{A},$$

where $\lambda_{h,a}(i)$ is the $i$-th element of $\lambda_{h,a}$ and $P_{a,i}$ is the $i$-th row of $P_a$.

Suppose that $(v^*, \lambda^*)$ is primal-dual optimal. As a result, it satisfies (a), (b), (c) and $v^*$ is the optimal value vector. By the definition of the optimal value function, we know that $v_h^*(i) - P_{\pi_h^*(i),i} v_{h+1}^* - r_{\pi_h^*(i)}(i)$. Since $\pi^*$ is unique, we have $v_h^*(i) > P_{a,i} v_{h+1}^* + r_a(i) = 0$ if $a \neq \pi_h^*(i)$. As a result, we have $\lambda_{h,a}^*(i) = 0$ for all $a \neq \pi_h^*(i)$.

We combine this observation with the dual feasibility relation $\sum_{a \in \mathcal{A}} \left(\lambda_{h,a} - P_a^T \lambda_{h-1,a}\right) = \xi_h$ and obtain

$$(\mathbf{I} - \Pi_{\pi^*}^T) \left(\lambda_{h,\pi_h^*(i)}^*(i)\right)_{h \in [H], i \in \mathcal{S}} = \xi,$$

where $\Pi_{\pi^*}$ is defined in Section 2.2. Note that $(\mathbf{I} - \Pi_{\pi^*}^T)$ is invertible as its determinant is equal to one. Then we have $\left(\lambda_{h,\pi_h^*(i)}^*(i)\right)_{h \in [H], i \in \mathcal{S}} = (\mathbf{I} - \Pi_{\pi^*}^T)^{-1}\xi$, which together with the complementarity condition implies that $\lambda^*$ is unique. Similarly we can show that $v^* = (\mathbf{I} - \Pi_{\pi^*})^{-1}(R_{h,\pi^*})_{h \in [H]}$ from the primal feasibility and slackness condition.

Now suppose that $(v^*, \lambda^*)$ satisfies the three conditions stated in Theorem 2. Then we obtain (a),(b),(c) directly, which proves that $(v^*, \lambda^*)$ is primal-dual optimal. ∎

**Saddle Point Formulation of Finite-Horizon MDP** According to the Lagrangian duality, we can rewrite the Bellman equation (8) into the following saddle point problem:

$$\min_{v \in \mathcal{V}} \max_{\lambda \in \Xi \cap \Delta} L(v, \lambda) = \sum_{h=0}^{H-1} \xi_h^T v_h + \sum_{h=0}^{H-1} \sum_{a \in \mathcal{A}} \lambda_{h,a}^T \left(r_a + P_a v_{h+1} - v_h\right), \quad (10)$$

where

$$\mathcal{V} = \left\{ v \mid v_h \geq 0, \|v_h\|_\infty \leq (H-h)\sigma, \forall\, h \in [H] \right\},$$

$$\Xi = \left\{ \lambda \mid \sum_{a \in \mathcal{A}} \lambda_{h,a} \geq \xi_h, \ \forall\, h \in [H] \right\}, \quad \Delta = \left\{ \lambda \mid \lambda \geq 0, \|\lambda_h\|_1 = \sum_{\tau=0}^{h} \|\xi_h\|_1, \forall\, h \in [H] \right\}. \tag{11}$$

We will prove later that the optimal primal solution $v^*$ and the optimal dual solutions $\lambda^*$ belong to the additional constraints $\mathcal{V}$ and $\Xi \cap \Delta$ respectively (Lemma 5). As a result, $(v^*, \lambda^*)$ is a pair of primal and dual solutions to (10).

We now state the matrix form of the saddle point problem. Let $\mathbf{I}$ be the identity matrix with dimension $|\mathcal{S}|H$ and let $\Pi_a \in \mathcal{R}^{|\mathcal{S}|H \times |\mathcal{S}|H}$ be the augmented transition probability matrix of the augmented Markov chain, taking the form

$$\Pi_a = \begin{bmatrix} 0 & P_a & 0 & \ldots & 0 \\ 0 & 0 & P_a & \ldots & 0 \\ \vdots & \vdots & \vdots & \ddots & \vdots \\ 0 & 0 & 0 & \ldots & P_a \\ 0 & 0 & 0 & \ldots & 0 \end{bmatrix}.$$

Then the Lagrangian $L(v, \lambda)$ is equivalent to

$$L(v, \lambda) = \xi^T v + \sum_{a \in \mathcal{A}} \lambda_a^T \left( R_a + (\Pi_a - \mathbf{I})v \right), \tag{12}$$

where $R_a = (r_a^T, \ldots, r_a^T)^T$, $\lambda_a = (\lambda_{h,a})_{h \in [H]}$. The partial derivatives of the Lagrangian are

$$\nabla_v L(v, \lambda) = \xi + \sum_{a \in \mathcal{A}} (\Pi_a^T - \mathbf{I})\lambda_a, \quad \nabla_{\lambda_a} L(v, \lambda) = R_a + (\Pi_a - \mathbf{I})v.$$

In what follows, we will utilize linear structures of the partial derivatives to develop stochastic primal-dual algorithms.

## 4. Stochastic Primal-Dual Methods for Reinforcement Learning

We are interested in developing algorithms that not only apply to explicitly given MDP models but also apply to reinforcement learning. In particular, we focus on the *model-free learning setting*, which is summarized as below.

**Model-Free Learning Setting of MDP**

(i) The state space $\mathcal{S}$, the action spaces $\mathcal{A}$, the reward upperbound $\sigma$, and the discount factor $\gamma$ (or the horizon $H$) are known.

(ii) The transition probabilities $\mathcal{P}$ and reward functions $r$ are unknown.

(iii) There is a Sampling Oracle ($\mathcal{SO}$) that takes input $(i, a)$ and generates a new state $j$ with probabilities $P_a(i, j)$ and a random reward $\hat{r}_{ija} \in [0, \sigma]$ with expectation $r_{ija}$.

Motivated by the value-policy duality (Theorems 1, 2), we develop a class of stochastic primal-dual methods for the saddle point formulation of Bellman Equation. In particular, we develop the Stochastic Primal-Dual method for Discounted Markov Decision Process (SPD-dMDP), as given in Algorithm1. We also develop the Stochastic Primal-Dual method for Finite-horizon Markov Decision Process (SPD-fMDP), as given by Algorithm 2. The SPD algorithms maintain an running estimate of the optimal value function (i.e., the primal solution) and the optimal policy (i.e, the dual solution). They make simple updates to the value and policy estimates as new state and reward observations are drawn from the sampling oracle.

---

**Algorithm 1** Stochastic Primal-Dual Algorithm for Discounted MDP (SPD-dMDP)

---

**Input:** Sampling Oracle $\mathcal{SO}$, $n = |\mathcal{S}|$, $m = |\mathcal{A}|$, $\gamma \in (0,1)$, $\sigma \in (0,\infty)$
Initialize $v^{(0)} : \mathcal{S} \mapsto \left[0, \frac{\sigma}{1-\gamma}\right]$ and $\lambda^{(0)} : \mathcal{S} \times \mathcal{A} \mapsto \left[0, \frac{\|\xi\|_1 \sigma}{1-\gamma}\right]$ arbitrarily.
Set $\xi = \frac{\sigma}{\sqrt{n}} e$
**for** $k = 1, 2, \ldots, T$ **do**
  Sample $i$ uniformly from $\mathcal{S}$
  Sample $a$ uniformly from $\mathcal{A}$
  Sample $j$ and $\hat{r}_{ija}$ conditioned on $(i,a)$ from $\mathcal{SO}$
  Set $\beta = \sqrt{n/k}$
  Update the primal iterates by

$$v^{(k)}(i) \leftarrow \max\left\{\min\left\{v^{(k-1)}(i) - \beta\left(\frac{1}{m}\xi(i) - \lambda_a^{(k-1)}(i)\right), \frac{\sigma}{1-\gamma}\right\}, 0\right\}$$

$$v^{(k)}(j) \leftarrow \max\left\{\min\left\{v^{(k-1)}(j) - \gamma\beta\lambda_a^{(k-1)}(i), \frac{\sigma}{1-\gamma}\right\}, 0\right\}$$

$$v^{(k)}(s) \leftarrow v^{(k-1)}(s) \quad \forall s \neq i, j$$

  Update the dual iterates by

$$\lambda_a^{(k-\frac{1}{2})}(i) \leftarrow \lambda^{(k-1)}(a,i) + \beta\left(\gamma v^{(k-1)}(j) - v^{(k-1)}(i) + \hat{r}_{ija}\right)$$

$$\lambda^{(k-\frac{1}{2})}(a', i') \leftarrow \lambda^{(k-1)}(a', i'), \quad \forall\, (a', i') \text{ such that } a' \neq a \text{ or } i' \neq i$$

  Project the dual iterates by

$$\lambda^{(k)} \leftarrow \Pi_{\Xi \cap \Delta} \lambda^{(k-\frac{1}{2})}, \text{ where } \Xi \text{ and } \Delta \text{ are given by (6)}$$

**end for**
**Ouput:** Averaged dual iterate $\hat{\lambda} = \frac{1}{T}\sum_{k=1}^T \lambda^{(k)}$ and randomized policy $\hat{\pi}$ where $\mathbf{P}(\hat{\pi}(i) = a) = \frac{\hat{\lambda}_a(i)}{\sum_{a \in \mathcal{A}} \hat{\lambda}_a(i)}$

---

**Implementation and Computational Complexities** The proposed SPD algorithms are essentially stochastic coordinate descent methods. They exhibit favorable properties such as small space complexity and small computational complexity per iteration.

The SPD-dMDP Algorithm 1 updates the value and policie estimates by processing one sample state-transition at a time. It keeps track of the dual variable $\lambda$ and primal variable $v$, which utilizes $|\mathcal{A}| \times |\mathcal{S}| + |\mathcal{S}| = \mathcal{O}(|\mathcal{A}| \times |\mathcal{S}|)$ space. In comparison, the dimension of the discounted MDP is $|\mathcal{A}|^2 \times |\mathcal{S}|$. Thus the space complexity of Algorithm 1 is sublinear with respect to the problem size. Moreover, the SPD-dMDP Algorithm 1 is a coordinate descent method. It updates two coordinates of the value estimation and one coordinate of the policy estimate per each iteration. It involves projection onto a special constraint, which is the intersection of the simplex and a box constraint. The projection is easy to compute and uses $\mathcal{O}(|\mathcal{S}| \times |\mathcal{A}|)$ arithmetic operations. The overall computation complexity per each iteration is $\mathcal{O}(|\mathcal{S}| \times |\mathcal{A}|)$ arithmetic operations. Therefore SPD Algorithm 1 uses sublinear space complexity and sublinear computation complexity per iteration. It is among one of the easiest to implement among existing reinforcement learning methods.

Algorithm 2 has a similar spirit with Algorithm 1. It keeps track of the dual variable $(\lambda_{h,a})_{h \in [H], a \in \mathcal{A}}$ (randomized policies for all periods) and primal variable $v_0, \ldots, v_{H-1}$ (value functions for all periods). Algorithm 2 is specially designed for $H$-period MDP in two aspects. It uses a non-uniform weight vector where $\xi_0 = \frac{e}{H}$ and $\xi_h = \frac{e}{(H-h)(H-h+1)}$, which places more weight for later periods to balance the smaller values. It uses larger stepsizes to update policies associated with later periods more aggressively, while using small stepsizes to update earlier-period policies more conservatively. The space complexity is $\mathcal{O}(|\mathcal{S}| \times |\mathcal{A}| \times H)$, which is sublinear with respect to the problem dimension $\mathcal{O}(|\mathcal{S}|^2 \times |\mathcal{A}| \times H)$. Algorithm 2 is essentially a coordinate descent method that involves projection onto simple sets. The computation complexity per each iteration is $\mathcal{O}(|\mathcal{S}| \times |\mathcal{A}| \times H)$, which is mainly due to the projection step.

**Comparisons with Existing Methods** Our SPD algorithms are fundamentally different from the existing reinforcement learning methods. From a theoretical perspective, our SPD algorithms are based on the stochastic saddle point formulation of Bellman equation. To authors' best knowledge, this idea has not been used in any existing method. From a practical perspective, the SPD methods are easy to implement and have small space and computational complexity (one of the smallest compared to existing methods). In what follows, we compare the newly proposed SDP methods with several popular existing methods.

- The new SPD methods share a similar spirit with the Q-learning and delayed Q-learning methods. Both of them maintain and update a value for each state-action pair $(i, a)$. Delayed Q-learning maintains estimates of the *value function* at each state-action pair $(i, a)$, i.e., the Q values. Our SPD maintains estimates of probabilities for choosing each $(i, a)$, i.e., the randomized policy. In both cases, the values associated with state-actions pairs are used to determine how to choose the actions. Delayed Q-learning uses $\mathcal{O}(|\mathcal{S}||\mathcal{A}|)$ space and $\mathcal{O}(\ln |\mathcal{A}|)$ arithmetic operations per iteration Strehl et al. (2006). Our SPD methods enjoy similar computational advantages as the delayed Q-learning method.

- The SPD methods are also related to the class of actor-critic methods. Our dual variable mimics the actor, while the primal variable mimics the critic. In particular, the dual update step in SPD turns out to be very similar to the actor update: both updates use a noisy temporal difference. Actor-critic methods are two-timescale

**Algorithm 2** Stochastic Primal-Dual Algorithm for Finite-horizon MDP (SPD-fMDP)

**Input:** Sampling Oracle $\mathcal{SO}$, $n = |\mathcal{S}|$, $m = |\mathcal{A}|$, $H$, $\sigma \in (0, \infty)$

Initialize $v_h : \mathcal{S} \mapsto [0, (H-h)\sigma]$ and $\lambda_h : \mathcal{S} \times \mathcal{A} \mapsto \left[0, \frac{n}{H-h}\right], \forall\, h \in [H]$ arbitrarily

Set $\xi_0 = \frac{e}{H}$ and $\xi_h = \frac{e}{(H-h)(H-h+1)}, \forall\, h \neq 0$

**for** $k = 1, 2, \ldots, T$ **do**

    Sample $i$ uniformly from $\mathcal{S}$

    Sample $a$ uniformly from $\mathcal{A}$

    Sample $j$ and $\hat{r}_{ija}$ conditioned on $(i,a)$ from $\mathcal{SO}$

    Update the primal iterates by

$$v_h^{(k)}(i) \leftarrow \max\left\{\min\left\{v_h^{(k-1)}(i) - \frac{(H-h)^2\sigma}{\sqrt{k}}\left(\xi_h(i) - m\lambda_{h,a}^{(k-1)}(i)\right),\quad (H-h)\sigma\right\},0\right\}, \forall\, h \in [H]$$

$$v_h^{(k)}(j) \leftarrow \max\left\{\min\left\{v_h^{(k-1)}(j) - \frac{m(H-h)^2\sigma}{\sqrt{k}}\lambda_{h-1,a}^{(k-1)}(i),\quad (H-h)\sigma\right\},0\right\}, \forall\, h \in [H]$$

$$v_h^{(k)}(s) \leftarrow v_h^{(k-1)}(s) \quad ,\forall\, h \in [H], s \neq i, j$$

    Update the dual iterates by

$$\lambda_{h,a}^{(k-\frac{1}{2})}(i) \leftarrow \lambda_{h,a}^{(k-1)}(i) + \frac{n}{(H-h)^2\sigma\sqrt{k}}\left(v_{h+1}^{(k-1)}(j) - v_h^{(k-1)}(i) + \hat{r}_{ija}\right), \forall\, h \in [H]$$

$$\lambda_{h,a'}^{(k-\frac{1}{2})}(i') \leftarrow \lambda_{h,a}^{(k-1)}(i), \quad \forall\, h \in [H], \forall\, a', i' \text{ such that } a' \neq a \text{ or } i' \neq i$$

    Project the dual iterates by

$$\lambda^{(k)} \leftarrow \Pi_{\Xi \cap \Delta}\lambda^{(k-\frac{1}{2})}, \text{ where } \Xi \text{ and } \Delta \text{ are given by (11)}$$

**end for**

**Ouput:** Averaged dual iterate $\hat{\lambda} = \frac{1}{T}\sum_{k=1}^{T}\lambda^{(k)}$ and randomized policy $\hat{\pi} = (\hat{\pi}_0, \cdots, \hat{\pi}_{H-1})$ where $\mathbf{P}(\hat{\pi}_h(i) = a) = \frac{\hat{\lambda}_{h,a}(i)}{\sum_{a \in \mathcal{A}} \hat{\lambda}_{h,a}(i)}$

---

methods in which the actor updates on a faster scale in comparison to the critic. In contrast, the new SPD methods have only one timescale: the primal and dual variables are updated using a single sequence of stepsizes. As a result, SPD methods are more efficient in utilizing new data as they emerge and achieve $\mathcal{O}(1/\sqrt{T})$ rate of convergence.

- Upper confidence methods maintain and update a value or interval for each state-action-state triplet; see the works by Lattimore and Hutter (2012), Dann and Brunskill (2015). These methods use space up to $\mathcal{O}(|\mathcal{S}|^2|\mathcal{A}|)$, which is linear with respect to the size of the MDP model. In contrast, SPD do not estimate transition probabilities of the unknown MDP and use only $\mathcal{O}(|\mathcal{S}||\mathcal{A}|)$ space.

To sum up, a main advantage of the SPD methods is the small storage and small computational complexity per iteration. We note that the main computational complexity of SPD is due to the projection of dual variables. It is possible to improve it to achieve $O(1)$ computational complexity per iteration by dropping the projection. This is to be improved in future research.

## 5. Main Results

In this section, we study the convergence of the two SPD methods: the SPD-dMDP Algorithm 1 and the SPD-fMDP Algorithm 2. Our main results show that the SPD methods output a randomized policy that is absolute-$\epsilon$-optimal using finite samples with high probability. We analyze the cases of discounted MDP and finite-horizon MDP separately.

### 5.1 Case I: Discounted-Reward Infinite-Horizon MDP

We analyze the SPD-dMDP Algorithm 1 as a stochastic analog of a deterministic primal-dual iteration. We show that the duality gap associated with the primal and dual iterates decreases to zero with the following guarantee.

**Theorem 3 (PAC Duality Gap)** *For any $\epsilon > 0$, $\delta \in (0,1)$, let $\hat{\lambda} = (\hat{\lambda}_a)_{a \in \mathcal{A}} \in \Re^{|\mathcal{S} \times \mathcal{A}|}$ be the averaged dual iterates generated by the SPD-dMDP Algorithm 1 using the following sample size/iteration number*

$$\Omega\left(\frac{|\mathcal{S}|^3 |\mathcal{A}|^2 \sigma^4}{(1-\gamma)^4 \epsilon^2} \ln\left(\frac{1}{\delta}\right)\right).$$

*Then the dual iterate $\hat{\lambda}$ satisfies*

$$\sum_{a \in \mathcal{A}} (\hat{\lambda}_a)^T (v^* - \gamma P_a v^* - r_a) \leq \epsilon$$

*with probability at least $1 - \delta$.*

*Proof Outline.* The SPD-dMDP Algorithm 1 can be viewed as a stochastic approximation scheme for the saddle point problem (5). Upon drawing a triplet $(i_k, a_k, j_k)$, we obtain noisy samples of partial derivatives $\nabla_v L(v^{(k)}, \lambda^{(k)})$ and $\nabla_\lambda L(v^{(k)}, \lambda^{(k)})$. the SPD-dMDP Algorithm 1 is equivalent to

$$v^{k+1} = \Pi_{\mathcal{V}}[v^k - \beta_k \left(\nabla_v L(v^{(k)}, \lambda^{(k)}) + \epsilon_k\right)]$$
$$\lambda^{k+1} = \Pi_{\Xi \cap \Delta}\left[\lambda^k + \beta_k \left(\nabla_\lambda L(v^{(k)}, \lambda^{(k)}) + \varepsilon_k\right)\right],$$

where $\mathcal{V}$, $\Xi$ and $\Delta$ are specially constructed sets, and $\epsilon_k, \varepsilon_k$ are zero-mean noise vectors. $\beta_k$ is the stepsize. By analyzing the distance between $(v^k, \lambda^k)$ and $(v^*, \lambda^*)$, we obtain that the squared distance decreases by factors of the duality gap per iteration. Then we construct a martingale based on the sequences of duality gaps and apply Bernstein's inequality. The formal proof is deferred to Section 6. ∎

Note that the dual variable is always nonnegative $\hat{\lambda} \geq 0$ by the projection on the nonnegative box $\mathcal{V}$. Also note that the nonnegative vector $v^* - (r_a + \gamma P_a v^*) \geq 0$ is a vector of primal constraint tolerances attained by the primal optimal solution $v^*$. Theorem 1 essentially gives an error bound for entries of $\hat{\lambda}$ corresponding to inactive primal row constraints.

Now we are ready to present the sample complexity of SPD for discounted MDP. Theorem 4 shows that the averaged dual iterate $\hat{\lambda}$ gives a randomized policy that approximates the optimal policy $\pi^*$. The performance of the randomized policy can be analyzed using the diminishing duality gap from Theorem 3.

**Theorem 4 (PAC Sample Complexity)** *For any $\epsilon > 0$, $\delta \in (0, 1)$, let the SPD-dMDP Algorithm 1 iterate with the following sample size/iteration number*

$$\Omega\left(\frac{|\mathcal{S}|^4|\mathcal{A}|^2\sigma^2}{(1-\gamma)^6\epsilon^2}\ln\left(\frac{1}{\delta}\right)\right),$$

*then the output policy $\hat{\pi}$ is absolute-$\epsilon$-optimal with probability at least $1-\delta$.*

Next we consider how to recover the optimal policy $\pi^*$ from the dual iterates $\hat{\lambda}$ generated by the SPD-dMDP Algorithm 1. Note that the policy space is discrete, which makes it possible to distinguish the optimal one from others when the estimated policy $\hat{\pi}$ is close enough to the optimal one.

**Definition 7** *Let the minimal action discrimination constant $\bar{d}$ be the minimal efficiency loss of deviating from the optimal policy $\pi^*$ by making a single wrong action. It is given by*

$$\bar{d} = \min_{(i,a):\pi^*(i)\neq a}(v^*(i) - \gamma P_{a,i}v^* - r_a(i)).$$

As long as there exists a unique optimal policy $\pi^*$, we have $\bar{d} > 0$. A large value of $\bar{d}$ means that it is easy to discriminate optimal actions from suboptimal actions. A small value of $\bar{d}$ means that some suboptimal actions perform similarly to optimal actions.

**Theorem 5 (Exact Recovery of The Optimal Policy)** *For any $\epsilon > 0$, $\delta \in (0, 1)$, let the SPD-dMDP Algorithm 1 iterate with the following sample size*

$$\Omega\left(\frac{|\mathcal{S}|^4|\mathcal{A}|^4\sigma^2}{\bar{d}^2(1-\gamma)^4}\ln\left(\frac{1}{\delta}\right)\right).$$

*Let $\hat{\pi}^{Tr}$ be obtained by rounding the randomized policy $\hat{\pi}$ to the nearest deterministic policy, given by*

$$\hat{\pi}^{Tr}(i) = \mathrm{argmax}_{a\in\mathcal{A}}\hat{\lambda}_{a,i}, \qquad i \in \mathcal{S}.$$

*Then $\mathbf{P}\left(\hat{\pi}^{Tr} = \pi^*\right) \geq 1 - \delta$.*

To our best knowledge, this is the first result that shows how to recover the exact optimal policy of reinforcement learning. The discrete nature of the policy space makes the exact recovery possible.

## 5.2 Case II: Finite-Horizon MDP

Now we analyze the SPD-fMDP Algorithm 2. Again we start with the duality gap analysis. We have the following theorem.

**Theorem 6 (PAC Duality Gap)** *For any $\epsilon > 0, \delta \in (0,1)$, let $\hat{\lambda} = (\hat{\lambda}_a)_{a \in \mathcal{A}} \in \Re^{|\mathcal{S} \times \mathcal{A}|}$ be the averaged dual iterates generated by the SPD-fMDP Algorithm 2 using the following sample size/iteration number*

$$\Omega\left(\frac{|\mathcal{S}|^4 |\mathcal{A}|^2 H^2 \sigma^2}{\epsilon^2}\left(\ln \frac{1}{\delta}\right)\right).$$

*Then the dual iterate $\hat{\lambda}$ satisfies*

$$\sum_{h=0}^{H-1} \sum_{a \in \mathcal{A}} (\hat{\lambda}_{h,a})^T (v_h^* - P_a v_{h+1}^* - r_a) \leq \epsilon.$$

*with probability at least $1 - \delta$.*

*Proof Outline.* We can view the SPD-fMDP Algorithm 2 as a stochastic approximation scheme for the saddle point problem (10). The SPD-fMDP Algorithm 2 is equivalent to the following iteration

$$v_h^{k+1} = \Pi_\mathcal{V}[v_h^k - \frac{(H-h)^2 \sigma \gamma_k}{n}\left(\nabla_{v_h} L(v^{(k)}, \lambda^{(k)}) + \epsilon_k\right)]$$

$$\lambda_h^{k+1} = \Pi_{\Xi \cap \Delta}\left[\lambda_h^k + \frac{\gamma_k}{m(H-h)^2 \sigma}\left(\nabla_{\lambda_h} L(v^{(k)}, \lambda^{(k)}) + \varepsilon_k\right)\right],$$

where $\Xi$ and $\Delta$ are given by (11), $\epsilon_k$ and $\varepsilon_k$ are zero-mean noise vectors, $\gamma_k$ is the stepsize. The formal proof is deferred to Section 6. ∎

Next we present the sample complexity for the SPD-fMDP Algorithm 2. The analysis is obtained from the duality gap.

**Theorem 7 (PAC Sample Complexity)** *For any $\epsilon > 0$, $\delta \in (0,1)$, let the SPD-fMDP Algorithm 2 iterate with the following sample size*

$$\Omega\left(\frac{|\mathcal{S}|^4 |\mathcal{A}|^2 H^6 \sigma^2}{\epsilon^2} \ln\left(\frac{1}{\delta}\right)\right),$$

*then the output policy $\hat{\pi}$ is absolute-$\epsilon$-optimal with probability at least $1 - \delta$.*

Next we consider how to recover the optimal policy $\pi^*$ from the dual iterates $\hat{\lambda}$ generated by the SPD-fMDP Algorithm 2. We abuse the notation $\bar{d}$ to denote the *minimal action discrimination* for finte-horizon MDP.

**Definition 8** *Let the minimal action discrimination constant $\bar{d}$ be the minimal efficiency loss of deviating from the optimal policy $\pi^*$ by making a single wrong action. It is given by*

$$\bar{d} = \min_{(h,i,a): \pi_h^*(i) \neq a} (v_h^*(i) - P_{a,i} v_{h+1}^* - r_a(i)).$$

As long as there exists a unique optimal policy $\pi^*$, we have $\bar{d} > 0$. Now we state our last theorem.

**Theorem 8 (Exact Recovery of The Optimal Policy)** *For any $\epsilon > 0$, $\delta \in (0,1)$, let $\hat{\pi}^{Tr}$ be the truncated pure policy such that*

$$\hat{\pi}_h^{Tr}(i) = argmax_{a \in \mathcal{A}} \hat{\lambda}_{h,a}(i), \qquad i \in \mathcal{S}_h.$$

*Let the SPD-fMDP Algorithm 2 iterate with the following iteration number/sample size*

$$\Omega\left(\frac{|\mathcal{S}|^4 |\mathcal{A}|^4 H^6 \sigma^2}{\bar{d}^2} \ln\left(\frac{1}{\delta}\right)\right).$$

*Then* $\mathbf{P}\left(\hat{\pi}^{Tr} = \pi^*\right) \geq 1 - \delta$.

The results of Theorems 6, 7, 8 for finite-horizon MDP are analogous to Theorems 3, 4, 5 for discounted-reward MDP. The horizon $H$ plays a role similar to the discounted infinite sum $\sum_{k=0}^{\infty} \gamma^k = \frac{1}{1-\gamma}$.

### 5.3 Summary, Discussions and Extensions

We have presented a novel primal-dual approach for solving the MDP in the model-free learning setting. A significant practical advantage of the primal-dual learning methods is the small storage and computational efficiency. The SPD methods use $\mathcal{O}(|\mathcal{S}| \times |\mathcal{A}|)$ space and $\mathcal{O}(|\mathcal{S}| \times |\mathcal{A}|)$ arithmetic operation per iteration, which is sublinear with respect to dimensions of the MDP. We show that the SPD output an absolute-$\epsilon$-optimal solution using $\mathcal{O}(\frac{|\mathcal{S}|^4 |\mathcal{A}|^2}{\epsilon^2})$ samples. In comparison, it is known that the sample complexity of reinforcement learning is bounded from below by $\mathcal{O}(\frac{|\mathcal{S}||\mathcal{A}|}{\epsilon^2})$ in a slightly different setting (Strehl et al., 2009), (Dann and Brunskill, 2015). Clearly, our sample complexity results do *not* yet match the lower bounds. The dependence on $\gamma, H, |\mathcal{S}|$ and $|\mathcal{A}|$ are to be improved.

We make several remarks about potential improvement and extensions of the primal-dual learning methods.

- The SPD-dMDP Algorithms 1 and 2 require that the state-action pair is sampled uniformly from $\mathcal{S} \times \mathcal{A}$. In other words, the SPD-dMDP Algorithm 1 and 2 use pure exploration without any exploitation. Such sampling oracle is suitable for off-line learning, when a fixed-size static data set is given. In the online learning setting, he sample complexity will be improved when actions are sampled according to the latest value and policy estimates rather than uniformly.

- A potential improvement is to drop or relax the constraint projection step of Algorithms 1,2. This would reduce the iteration complexity from $\mathcal{O}(|\mathcal{S}||\mathcal{A}|)$ to $\mathcal{O}(1)$, making each iteration a real coordinate descent iteration. However, this would impact the convergence analysis and require additional mechanism to ensure the boundedness of iterates. This is left for future research.

- Another extension is to consider average-reward discounted MDP. In the average-reward case, the discount factor $\gamma$ and horizon $H$ will disappear from the sample complexity. One cannot derive the sample complexity for average-reward MDP from the existing result. We conjecture that the sample complexity of average-reward MDP depends on the spectral gap of the underlying Markov chains.

To the authors' belief, the primal-dual approach studied in this paper has significant theoretical and practical potential. The bilinear stochastic saddle point formulation of Bellman equations is amenable to online learning and dimension reduction. The intrinsic linear duality between the optimal policy and value functions implies a convenient structure for efficient learning.

## 6. Technical Proofs

In this section, we provide the proofs to the theorems in Section 5. For readers who are not concerned with the technical details, this part can be safely skipped.

### 6.1 Analysis of the SPD-dMDP Algorithm 1

In what follows, we denote $n = |\mathcal{S}|$ and $m = |\mathcal{A}|$ for short. For any $i, j \in \mathcal{S}$, we denote $E_{i_k, j_k}$ the matrix with $(i, j)$th entry being 1 and all other entries being zeros. Then the SPD-dMDP Algorithm 1 is a special case of the following iteration:

$$v^{k+1/2} = v^k - \beta_{k+1}\big(\frac{1}{m}\xi_{i_k}e_{i_k} + (\gamma E_{j_k, i_k} - E_{i_k, i_k})\lambda^k_{a_k}\big)$$

$$\lambda^{k+1/2}_a = \begin{cases} \lambda^k_a + \beta_{k+1}\big((\gamma E_{i_k, j_k} - E_{i_k, i_k})v^k + r_{i_k j_k a_k}e_{i_k}\big) & \text{if } a = a^k, \\ \lambda^k_a & \text{if } a \neq a^k, \end{cases}$$

$$v^{k+1} = \Pi_\mathcal{V} v^{k+1/2},$$

$$\lambda^{k+1} = \Pi_{\Xi \cap \Delta} \lambda^{k+1/2}.$$

In what follows, we first analyze the sequence of duality gaps associated with the iterates. Then we construct a martingale based on the error sequence and duality gaps and provide an error bounds using the Bernstein inequality. Finally we show that the duality gap gives an upper bound on the sub-optimality of the output randomized policy. We denote by $\mathscr{F}_k$ the collection of all random variables generated up to iteration $k$.

#### 6.1.1 Preliminary Lemmas

**Lemma 1** *Suppose that $(v^*, \lambda^*)$ is a pair of primal and dual solutions to the linear programs* (2),(3). *Then*

$$\|v^*\|_\infty \leq \frac{\sigma}{1-\gamma}, \qquad \|v^*\|_2 \leq \frac{\sigma\sqrt{n}}{1-\gamma}, \qquad \|\lambda^*\|_2 \leq \|\lambda^*\|_1 = \frac{\|\xi\|_1}{1-\gamma}, \qquad \xi \leq \sum_{a \in \mathcal{A}} \lambda^*_a.$$

*Proof.* (i) Note that $v^* = r_{\pi^*} + \gamma P_{\pi^*} v^*$. We have

$$\|v^*\|_\infty \leq \|r_{\pi^*}\|_\infty + \gamma \|P_{\pi^*}\|_\infty \|v^*\|_\infty \leq \|r_{\pi^*}\|_\infty + \gamma \|v^*\|_\infty.$$

Then we have

$$\|v^*\|_\infty \leq \frac{\|r_{\pi^*}\|_\infty}{1-\gamma} \leq \frac{\sigma}{1-\gamma}.$$

(ii) We have $\|v^*\|_2 \leq \sqrt{n}\|v^*\|_\infty \leq \frac{\sqrt{n}\sigma}{1-\gamma}$.

(iii) Similarly we note that $\lambda^* = \xi + \gamma(P_{\pi^*})^T\lambda^*$. Noting that $\lambda^* \geq 0$ and $\xi \geq 0$, we have

$$\|\lambda^*\|_1 = e^T\lambda^* = e^T\xi + \gamma e^T(P_{\pi^*})^T\lambda^* = e^T\xi + \gamma e^T\lambda^* = \|\xi\|_1 + \gamma\|\lambda^*\|_1.$$

Therefore we have $(1-\gamma)\|\lambda^*\|_1 = \|\xi\|_1$ so that $\|\lambda^*\|_2 \leq \|\lambda^*\|_1 = \frac{\|\xi\|_1}{1-\gamma}$.

(iv) We use the dual feasibility constraint and obtain

$$(\sum_{a\in\mathcal{A}} \lambda_a^*)(\mathbf{I} - \gamma P_{\pi^*}^T) = \sum_{a\in\mathcal{A}} \lambda_a^*(\mathbf{I} - \gamma P_a^T) = \xi,$$

which implies $\sum_{a\in\mathcal{A}} \lambda_a^* \geq \xi$ elementwise. ∎

**Lemma 2** *The solution to the modified saddle point problem (5) is equal to the solution of the original problem (4).*

*Proof.* The proof is straightforward by using Lemma 1 with the definition of $\mathcal{V}, \Delta$.

**Lemma 3** *We denote for short that*

$$\mathcal{E}_k = \|v^k - v^*\|^2 + \sum_{a\in\mathcal{A}} \|\lambda_a^k - \lambda_a^*\|^2, \qquad \mathcal{G}_k = \sum_{a\in\mathcal{A}} (\lambda_a^k)^T(v^* - \gamma P_a v^* - r_a).$$

*Then the SPD-dMDP Algorithm 1 satisfies for all $k$ with probability 1 that*

$$\mathcal{E}_k \leq D^2, \qquad \mathbf{E}[\mathcal{E}_{k+1} \mid \mathscr{F}_k] \leq \mathcal{E}_k - \frac{2\beta_0/(mn)}{\sqrt{k+1}}\mathcal{G}_k + \frac{\beta_0^2}{k+1}B^2, \tag{13}$$

*where $D^2 = \frac{n\sigma^2}{(1-\gamma)^2} + \frac{\|\xi\|_1^2}{(1-\gamma)^2}$, $B^2 = \mathcal{O}\left(\frac{\|\xi\|_1^2/(mn)+\sigma^2}{(1-\gamma)^2}\right)$.*

*Proof.* We can apply Lemma 1 and get

$$\mathcal{E}_k \leq D^2 := \frac{n\sigma^2}{(1-\gamma)^2} + \frac{\|\xi\|_1^2}{(1-\gamma)^2}, \qquad \forall k.$$

We use the nonexpansiveness of $\Pi_{\Xi\cap\Delta}$ and the facts $\lambda^* \in \Xi, \lambda^* \in \Delta$, and we derive

$$\begin{aligned}
&\|v^{k+1} - v^*\|^2 + \sum_{a\in\mathcal{A}}\|\lambda_a^{k+1} - \lambda_a^*\|^2 \\
\leq& \|v^{k+1/2} - v^*\|^2 + \sum_{a\in\mathcal{A}}\|\lambda_a^{k+1/2} - \lambda_a^*\|^2 \\
\leq& \|v^k - v^*\|^2 + \sum_{a\in\mathcal{A}}\|\lambda_a^k - \lambda_a^*\|^2 - 2\beta_{k+1}\underbrace{\left(\frac{1}{m}\xi_{i_k}e_{i_k} + (\gamma E_{j_k,i_k} - E_{i_k,i_k})\lambda_{a_k}^k\right)^T(v^k - v^*)}_{\Psi_1} \\
&+ 2\beta_{k+1}\underbrace{\left((\gamma E_{i_k,j_k} - E_{i_k,i_k})v^k + \hat{r}_{a_k,i_k}e_{i_k}\right)^T(\lambda_{a_k}^k - \lambda_{a_k}^*)}_{\Psi_2} \\
&+ \beta_{k+1}^2\underbrace{\left\|\frac{1}{m}\xi_{i_k}e_{i_k} + (\gamma E_{j_k,i_k} - E_{i_k,i_k})\lambda_{a_k}^k\right\|^2}_{\Phi_1} + \beta_{k+1}^2\underbrace{\|(\gamma E_{i_k,j_k} - E_{i_k,i_k})v^k + \hat{r}_{a_k,i_k}e_i\|^2}_{\Phi_2}.
\end{aligned}$$

(14)

In what follows, we analyze the righthandside of the preceding inequality.

1. We analyze the term $\Psi_1$. We have

$$\begin{aligned}
\mathbf{E}\left[\Psi_1 \mid \mathscr{F}_k\right] &= \mathbf{E}\left[\left(\frac{1}{m}\xi_{i_k}e_{i_k} + (\gamma E_{j_k,i_k} - E_{i_k,i_k})\lambda_{a_k}^k\right) \Big| \mathscr{F}_k\right]^T (v^k - v^*) \\
&= \frac{1}{mn}\xi^T(v^k - v^*) + \frac{1}{m}\sum_{a\in\mathcal{A}}(\lambda_a^k)^T\mathbf{E}\left[(\gamma E_{j_k,i_k} - E_{i_k,i_k}) \mid a_k = a, \mathscr{F}_k\right]^T(v^k - v^*) \\
&= \frac{1}{mn}\xi^T(v^k - v^*) + \frac{1}{mn}\sum_{a\in\mathcal{A}}(\lambda_a^k)^T(\gamma P_a - \mathbf{I})(v^k - v^*) \\
&= \frac{1}{mn}\sum_{a\in\mathcal{A}}(\lambda_a^k - \lambda_a^*)^T(\gamma P_a - \mathbf{I})(v^k - v^*),
\end{aligned}$$

where the last relation uses the fact $\sum_{a\in\mathcal{A}}(\mathbf{I} - \gamma P_a^T)\lambda_a^* = \xi$ (Theorem 1).

2. We analyze the term $\Psi_2$.

$$\begin{aligned}
\mathbf{E}\left[\Psi_2 \mid \mathscr{F}_k\right] &= \mathbf{E}\left[\left((\gamma E_{i_k,j_k} - E_{i_k,i_k})v^k + \hat{r}_{a_k,i_k}e_{i_k}\right)^T(\lambda_{a_k}^k - \lambda_{a_k}^*) \mid \mathscr{F}_k\right] \\
&= \frac{1}{mn}\sum_{a\in\mathcal{A}}((\gamma P_a - \mathbf{I})v^k + r_a)^T(\lambda_a^k - \lambda_a^*) \\
&= \frac{1}{mn}\sum_{a\in\mathcal{A}}(\lambda_a^k - \lambda_a^*)^T((\gamma P_a - \mathbf{I})v^k + r_a)
\end{aligned}$$

Combining the preceding relations, we obtain

$$\mathbf{E}\left[\Psi_1 - \Psi_2 \mid \mathscr{F}_k\right] = \frac{1}{mn}\sum_{a\in\mathcal{A}}(\lambda_a^k)^T(v^* - \gamma P_a v^* - r_a).$$

3. Next, we analyze $\mathbf{E}[\Phi_1 \mid \mathscr{F}_k]$. Note that

$$\begin{aligned}
\mathbf{E}[\Phi_1 \mid \mathscr{F}_k] &= \mathbf{E}\left[\left\|\frac{1}{m}\xi_{i_k}e_{i_k} + (\gamma E_{j_k,i_k} - E_{i_k,i_k})\lambda_{a_k}^k\right\|^2 \Big| \mathscr{F}_k\right] \\
&\leq \frac{2}{m^2n}\xi^T\xi + 2\mathbf{E}[\|(\gamma E_{j_k,i_k} - E_{i_k,i_k})\lambda_{a_k}^k\|^2 \mid \mathscr{F}_k] \\
&\leq \frac{2}{m^2n}\xi^T\xi + 2\mathbf{E}[(1+\gamma)^2(\lambda_{a_k,i_k}^k)^2 \mid \mathscr{F}_k] \qquad (15)\\
&\leq \frac{2\|\xi\|_2^2}{m^2n} + \frac{8}{mn}\sum_{a\in\mathcal{A}}\|\lambda_a^k\|^2 \\
&\leq \frac{2\|\xi\|_2^2}{m^2n} + \frac{8\|\xi\|_1^2/(mn)}{(1-\gamma)^2},
\end{aligned}$$

where the last relation uses Lemma 1.

4. Similarly, we analyze $\mathbf{E}[\Phi_2|\mathscr{F}_k]$ and obtain

$$\begin{aligned}
\mathbf{E}\left[\Phi_2 \mid \mathscr{F}_k\right] &= \mathbf{E}\left[\|(\gamma E_{i_k,j_k} - E_{i_k,i_k})v^k + \hat{r}_{a_k,i_k}e_{i_k}\|^2 \mid \mathscr{F}_k\right] \\
&= \mathbf{E}\left[\left(\gamma v^k(j_k) - v^k(i_k) + \hat{r}_{a_k,i_k}\right)^2 \mid \mathscr{F}_k\right] \\
&\leq 2\mathbf{E}\left[\left(\gamma v^k(j_k) - v^k(i_k)\right)^2 \mid \mathscr{F}_k\right] + 2\mathbf{E}\left[(\hat{r}_{a_k,i_k})^2 \mid \mathscr{F}_k\right] \quad (16) \\
&\leq 8\|v^k\|_\infty^2 + 2\sigma^2 \\
&\leq \frac{10\sigma^2}{(1-\gamma)^2}.
\end{aligned}$$

In what follows, we analyze the main iteration. We use the notation

$$\mathcal{E}_k = \|v^k - v^*\|^2 + \sum_{a\in\mathcal{A}}\|\lambda_a^k - \lambda_a^*\|^2, \qquad \mathcal{G}_k = \sum_{a\in\mathcal{A}}(\lambda_a^k)^T(v^* - \gamma P_a v^* - r_a).$$

Taking conditional expectation on (14) and substituting all the above equations, we have

$$\mathbf{E}\left[\mathcal{E}_{k+1} \mid \mathscr{F}_k\right] \leq \mathcal{E}_k - \beta_{k+1}\frac{2}{mn}\mathcal{G}_k + \beta_{k+1}^2\left(\mathbf{E}[\Phi_1|\mathscr{F}_k] + \mathbf{E}[\Phi_2|\mathscr{F}_k]\right). \quad (17)$$

We apply equations (15), (16) and let $\beta_k = \frac{\beta_0}{\sqrt{k}}$. It follows that

$$\mathbf{E}\left[\mathcal{E}_{k+1} \mid \mathscr{F}_k\right] \leq \mathcal{E}_k - \frac{2\beta_0/(mn)}{\sqrt{k+1}}\mathcal{G}_k + \frac{\beta_0^2}{k+1}\mathcal{O}\left(\frac{\|\xi\|_1^2/(mn) + \sigma^2}{(1-\gamma)^2}\right),$$

for all $k$ with probability 1. ∎

**Lemma 4** *Let $D, B$ be the constants given in Lemma 3. Then*

$$\frac{\sqrt{k+1}}{2\beta_0}\left(\mathbf{E}\left[(\mathcal{E}_{k+1} - \mathbf{E}\left[\mathcal{E}_{k+1} \mid \mathscr{F}_k\right])^2\right]\right)^{1/2} \leq \mathcal{O}(DB),$$

*and there exists $M > 0$ such that*

$$\frac{\sqrt{k+1}}{2\beta_0}\left(\mathcal{E}_{k+1} - \mathbf{E}\left[\mathcal{E}_{k+1} \mid \mathscr{F}_k\right]\right) \leq M, \quad w.p.\ 1.$$

*Proof.* To see this, we obtain that

$$\mathcal{E}_{k+1} - \mathcal{E}_k = \|v^{k+1} - v^*\|^2 - \|v^k - v^*\|^2 + \sum_{a\in\mathcal{A}}\|\lambda_a^{k+1} - \lambda_a^*\|^2 - \sum_{a\in\mathcal{A}}\|\lambda_a^k - \lambda_a^*\|^2$$
$$= 2(v^{k+1} - v^k)^T(v^k - v^*) + \|v^{k+1} - v^k\|^2 + 2(\lambda^{k+1} - \lambda^k)^T(\lambda^k - \lambda^*) + \|\lambda^{k+1} - \lambda^k\|^2.$$

By using basic inequalities, we have

$$|\mathcal{E}_{k+1} - \mathcal{E}_k| \leq 4D_v\|v^{k+1} - v^k\| + \|v^{k+1} - v^k\|^2 + 4D_\lambda\|\lambda^{k+1} - \lambda^k\| + \|\lambda^{k+1} - \lambda^k\|^2,$$

where we denote for short that $D_v^2 = \frac{n\sigma^2}{(1-\gamma)^2}, D_\lambda^2 = \frac{\|\xi\|_1^2}{(1-\gamma)^2}$ and they satisfy $D^2 = D_v^2 + D_\lambda^2$. The inequality is due to the observation that $(v^{k+1} - v^k)^T(v^k - v^*) \leq \|v^{k+1} - v^k\|\|v^k - v^*\|$. Note that by the nonexpansiveness of projection, we have

$$\mathbf{E}\left[\|v^{k+1} - v^k\|^2 \mid \mathscr{F}_k\right] \leq \mathbf{E}\left[\|v^{k+1/2} - v^k\|^2 \mid \mathscr{F}_k\right] \leq \beta_k^2 \mathbf{E}\left[\Phi_1 \mid \mathscr{F}_k\right],$$

and

$$\mathbf{E}\left[\|\lambda^{k+1} - \lambda^k\|^2 \mid \mathscr{F}_k\right] \leq \mathbf{E}\left[\|\lambda^{k+1/2} - \lambda^k\|^2 \mid \mathscr{F}_k\right] \leq \beta_k^2 \mathbf{E}\left[\Phi_2 \mid \mathscr{F}_k\right].$$

Then

$$\mathbf{E}\left[|\mathcal{E}_{k+1} - \mathcal{E}_k|^2 \mid \mathscr{F}_k\right] \leq 32\beta_k^2(D_v^2\mathbf{E}\left[\Phi_1 \mid \mathscr{F}_k\right] + D_\lambda^2\mathbf{E}\left[\Phi_2 \mid \mathscr{F}_k\right]) + O(\beta_k^4) = \mathcal{O}(\beta_k^2 D^2 B^2).$$

Note that $\mathcal{E}_k \in \mathscr{F}_k$, therefore $\mathbf{E}\left[\mathcal{E}_k \mid \mathscr{F}_k\right] = \mathcal{E}_k$ and

$$\mathcal{E}_{k+1} - \mathbf{E}\left[\mathcal{E}_{k+1} \mid \mathscr{F}_k\right] = \mathcal{E}_{k+1} - \mathcal{E}_k - \mathbf{E}\left[\mathcal{E}_{k+1} - \mathcal{E}_k \mid \mathscr{F}_k\right].$$

Then we have

$$\begin{aligned}\mathbf{E}\left[|\mathcal{E}_{k+1} - \mathbf{E}\left[\mathcal{E}_{k+1} \mid \mathscr{F}_k\right]|^2\right] &= \mathbf{E}\left[|\mathcal{E}_{k+1} - \mathcal{E}_k - \mathbf{E}\left[\mathcal{E}_{k+1} - \mathcal{E}_k \mid \mathscr{F}_k\right]|^2\right] \\ &= \mathbf{E}\left[\mathbf{E}\left[|\mathcal{E}_{k+1} - \mathcal{E}_k - \mathbf{E}\left[\mathcal{E}_{k+1} - \mathcal{E}_k \mid \mathscr{F}_k\right]|^2 \mid \mathscr{F}_k\right]\right] \\ &\leq \mathbf{E}\left[\mathbf{E}\left[|\mathcal{E}_{k+1} - \mathcal{E}_k|^2 \mid \mathscr{F}_k\right]\right] \\ &\leq \mathcal{O}(\beta_k^2 D^2 B^2),\end{aligned}$$

where the inequality uses the fact that the variance of a random variable is bounded by its second moment. In the second equality, the first expectation is taken over $\mathscr{F}_k$ and the second expectation is taken over all the randomness at round $k + 1$.

We then prove the second statement. Denote $\Delta v^k = \frac{1}{m}\xi_{i_k}e_{i_k} + (\gamma E_{j_k,i_k} - E_{i_k,i_k})\lambda_{a_k}^k$. Then we have

$$\|v^{k+1} - v^*\|^2 - \mathbf{E}\left[\|v^{k+1} - v^*\|^2 \mid \mathscr{F}_k\right] \leq \|v^k - v^* - \beta_k\Delta v^k\|^2 - \mathbf{E}\left[\|v^k - v^* - \beta_k\Delta v^k\|^2 \mid \mathscr{F}_k\right]$$
$$= \mathcal{O}(\beta_k\|v^k - v^*\|_2\|\Delta v^k\|_2) \leq \mathcal{O}\left(\frac{\beta_0\sigma\sqrt{n}\|\xi\|_1}{\sqrt{k}(1-\gamma)^2}\right),$$

where the last inequality is due to Lemma 1. Therefore, $\frac{\sqrt{k+1}}{2}\left(\|v^{k+1} - v^*\|^2 - \mathbf{E}\left[\|v^{k+1} - v^*\|^2 \mid \mathscr{F}_k\right]\right)$ is smaller than $M$ up to a multiplicative constant where $M = \frac{\sigma\beta_0\sqrt{n}\|\xi\|_1}{(1-\gamma)^2}$. Similarly, we can prove that $\frac{\sqrt{k+1}}{2}\sum_{a \in \mathcal{A}}\|\lambda_a^{k+1} - \lambda_a^*\|^2 - \mathbf{E}\left[\|\lambda_a^{k+1} - \lambda_a^*\|^2 \mid \mathscr{F}_k\right]$ is smaller than $M$ to a multiplicative constant. To this point, we have proved the lemma. ∎

### 6.1.2 Proofs of Theorems 3, 4, and 5

**Proof of Theorem 3** Rearranging the terms of (13), we have

$$\frac{1}{mn}\mathcal{G}_k \leq \frac{\sqrt{k+1}}{2\beta_0}(\mathcal{E}_k - \mathbf{E}\left[\mathcal{E}_{k+1} \mid \mathscr{F}_k\right]) + \frac{\beta_0}{2\sqrt{k+1}}B^2,$$

where we denote $B^2 = \mathcal{O}\left(\frac{\|\xi\|_1^2/(mn)+\sigma^2}{(1-\gamma)^2}\right)$ for short.

Summing over $k$ and taking average, we have

$$\frac{1}{T}\frac{1}{mn}\sum_{k=1}^T \mathcal{G}_k$$

$$\leq \frac{1}{\beta_0 T}\sum_{k=1}^T \frac{\sqrt{k+1}}{2}(\mathcal{E}_k - \mathbf{E}\left[\mathcal{E}_{k+1} \mid \mathscr{F}_k\right]) + \frac{1}{2T}\sum_{k=1}^T \frac{\beta_0}{\sqrt{k+1}}B^2$$

$$\leq \frac{1}{\beta_0 T}\sum_{k=1}^T \frac{\sqrt{k+1}-\sqrt{k}}{2}\mathcal{E}_k + \frac{1}{\beta_0 T}\sum_{k=1}^T \frac{\sqrt{k+1}}{2}\left(\mathcal{E}_{k+1} - \mathbf{E}\left[\mathcal{E}_{k+1} \mid \mathscr{F}_k\right]\right) + \frac{1}{2T\beta_0}\mathcal{E}_1 + \mathcal{O}\left(\frac{\beta_0 B^2}{\sqrt{T}}\right).$$

Let us construct a sequence of random variables $\{M_t\}$ given by

$$M_{t+1} = \sum_{k=1}^t \frac{\sqrt{k+1}}{2}\left(\mathcal{E}_{k+1} - \mathbf{E}\left[\mathcal{E}_{k+1} \mid \mathscr{F}_k\right]\right).$$

By the construction of $M_t$, we have $\mathbf{E}\left[M_{t+1} \mid \mathscr{F}_t\right] = M_t$, which implies that $M_t$ is a martingale. According to Lemma 4, we also have $\frac{\sqrt{k+1}}{2}\left(\mathcal{E}_{k+1} - \mathbf{E}\left[\mathcal{E}_{k+1} \mid \mathscr{F}_k\right]\right) \leq M$ and $\frac{\sqrt{k+1}}{2\beta_0}\left(\mathbf{E}\left[(\mathcal{E}_{k+1} - \mathbf{E}\left[\mathcal{E}_{k+1} \mid \mathscr{F}_k\right])^2\right]\right)^{1/2} \leq \mathcal{O}(DB)$ with probability 1. As a result $M_t$ is a martingale with bounded difference and its difference has bounded second moment.

Let $\epsilon, \delta$ be arbitrarily given scalers, we apply the Bernstein inequality and obtain that

$$\mathbf{P}\left(\frac{1}{T}M_T \geq \frac{\beta_0 \epsilon}{mn}\right) \leq \exp\left(-\frac{T(\frac{\beta_0\epsilon}{mn})^2}{2(BD\beta_0)^2 + (2/3)M(\frac{\beta_0\epsilon}{mn})}\right).$$

By taking $T \geq 2\max\left\{\frac{2(BD)^2 m^2 n^2 \ln(1/\delta)}{\epsilon^2}, \frac{2Mmn\ln(1/\delta)}{3\beta_0\epsilon}\right\}$, we obtain $\mathbf{P}\left(\frac{1}{T}\sum_{k=1}^T M_k \geq \frac{\beta_0\epsilon}{mn}\right) \leq \delta$. Then with probability at least $1 - \delta$, we have

$$\frac{1}{T}\frac{1}{mn}\sum_{k=1}^T \mathcal{G}_k \leq \frac{1}{2\beta_0\sqrt{T}}D^2 + \frac{\epsilon}{mn} + \mathcal{O}\left(\frac{\beta_0 B^2}{\sqrt{T}}\right).$$

By taking $\beta_0 = \sqrt{n}$ and $\|\xi\|_1^2 = n\sigma^2$, we obtain

$$\frac{1}{T}\sum_{k=1}^T \mathcal{G}_k = \mathcal{O}\left(\frac{mnDB}{\sqrt{T}}\right) + \epsilon = \mathcal{O}\left(\frac{n^{3/2}m\sigma^2}{(1-\gamma)^2\sqrt{T}}\right) + \epsilon,$$

with probability at least $1 - \delta$. When $\epsilon < \frac{3\beta_0 mn(BD)^2}{M}$, we have $T \geq \frac{4(BD)^2 m^2 n^2 \ln(1/\delta)}{\epsilon^2} = \frac{4m^2n^3\sigma^4}{(1-\gamma)^4\epsilon^2}\ln\left(\frac{1}{\delta}\right)$. Plugging this relationship into the preceding equality, we have $\frac{1}{T}\sum_{k=1}^T \mathcal{G}_k = \mathcal{O}(\epsilon)$ with probability $1 - \delta$. ∎

**Proof of Theorem 4** In the following, we show that the duality gap for the bilinear saddle point problem gives a bound on the efficiency loss of the randomized policy $\hat{\pi}$. Note that $\hat{\lambda} = \frac{1}{T}\sum_{k=1}^{T} \lambda^k$. So we have

$$\left(\frac{1}{T}\sum_{k=1}^{T} \mathcal{G}_k\right) = \sum_{a \in \mathcal{A}} (\hat{\lambda}_a)^T (v^* - \gamma P_a v^* - r_a)$$

$$= \sum_{a \in \mathcal{A}, i \in \mathcal{S}} \hat{\lambda}_{a,i}(v_i^* - \gamma P_{a,i} v^* - r_{a,i})$$

$$= \sum_{i \in \mathcal{S}} \left(\sum_{a' \in \mathcal{A}} \hat{\lambda}_{a',i}\right) \sum_{a \in \mathcal{A}} \frac{\hat{\lambda}_{a,i}}{\sum_{a' \in \mathcal{A}} \hat{\lambda}_{a',i}} (v_i^* - \gamma P_{a,i} v^* - r_{a,i}).$$

We denote

$$r_i^{\hat{\pi}} = \sum_{a \in \mathcal{A}} \frac{\hat{\lambda}_{a,i}}{\sum_{a' \in \mathcal{A}} \hat{\lambda}_{a',i}} r_{a,i}, \qquad P_i^{\hat{\pi}} = \sum_{a \in \mathcal{A}} \frac{\hat{\lambda}_{a,i}}{\sum_{a' \in \mathcal{A}} \hat{\lambda}_{a',i}} P_{a,i}.$$

We observe that $r^{\hat{\pi}}$ and $P^{\hat{\pi}}$ are the expected state transition reward and transition probability matrix under the randomized policy $\hat{\pi}$. Note that the value vector $v^{\hat{\pi}}$ under policy $\hat{\pi}$ satisfies

$$r^{\hat{\pi}} = v^{\hat{\pi}} - \gamma P^{\hat{\pi}} v^{\hat{\pi}}.$$

It follows that

$$\left(\frac{1}{T}\sum_{k=1}^{T} \mathcal{G}_k\right) = \sum_{i \in \mathcal{S}} \left(\sum_{a' \in \mathcal{A}} \hat{\lambda}_{a',i}\right) (v_i^* - \gamma P_i^{\hat{\pi}} v^* - r_i^{\hat{\pi}})$$

$$= \sum_{i \in \mathcal{S}} \left(\sum_{a' \in \mathcal{A}} \hat{\lambda}_{a',i}\right) (e_i - \gamma P_i^{\hat{\pi}})(v^* - v^{\hat{\pi}})$$

$$= \left(\sum_{a \in \mathcal{A}} \hat{\lambda}_a\right)^T (\mathbf{I} - \gamma P^{\hat{\pi}})(v^* - v^{\hat{\pi}})$$

Note that $\left(\sum_{a \in \mathcal{A}} \hat{\lambda}_{a,i}\right) \geq \xi_i$ due to the constraint projection step of the SPD-dMDP Algorithm 1. Also note that $v_i^* - (r_i^{\hat{\pi}} + \gamma P_i^{\hat{\pi}} v^*) \geq 0$ for all $i$ and $a$ due to the fact that $v^*$ is the value corresponding to the optimal policy. As a result, we have $(\mathbf{I} - \gamma P^{\hat{\pi}})(v^* - v^{\hat{\pi}}) \geq 0$. Then we have

$$\sum_{a \in \mathcal{A}} (\hat{\lambda}_a)^T (v^* - r_a - \gamma P_a v^*) \geq \xi^T (\mathbf{I} - \gamma P^{\hat{\pi}})(v^* - v^{\hat{\pi}}) \geq \left(\min_{i \in \mathcal{S}} \xi_i\right) \|(\mathbf{I} - \gamma P^{\hat{\pi}})(v^{\hat{\pi}} - v^*)\|_\infty.$$

We have

$$\|(\mathbf{I} - \gamma P^{\hat{\pi}})(v^{\hat{\pi}} - v^*)\|_\infty \geq \|v^{\hat{\pi}} - v^*\|_\infty - \|\gamma P^{\hat{\pi}}(v^{\hat{\pi}} - v^*)\|_\infty$$

$$\geq \|v^{\hat{\pi}} - v^*\|_\infty - \|\gamma P^{\hat{\pi}}\|_\infty \|(v^{\hat{\pi}} - v^*)\|_\infty$$

$$= (1 - \gamma)\|v^{\hat{\pi}} - v^*\|_\infty.$$

For any $\epsilon_1 > 0, 0 \leq \delta < 1$, taking $\xi = \frac{\sigma}{\sqrt{n}} e$ and applying Theorem 3 with $\epsilon = \epsilon_1 \cdot (1-\gamma) \cdot \frac{\sigma}{\sqrt{n}}$ gives us the following: if $T = \Omega\left(\frac{|\mathcal{S}|^4|\mathcal{A}|^2\sigma^2}{(1-\gamma)^6\epsilon^2} \ln\left(\frac{1}{\delta}\right)\right)$, then

$$\|v^{\hat{\pi}} - v^*\|_\infty \leq \frac{1}{(1-\gamma)\min_i \xi_i} \left(\frac{1}{T}\sum_{k=1}^{T} \mathcal{G}_k\right) = \epsilon_1,$$

with probability at least $1-\delta$. ∎

**Proof of Theorem 5** We denote for short that $d_{i,a} = v_i^* - (\gamma P_{a,i} v^* + r_{a,i})$. By using the complementarity condition, we have $d_{i,a} > 0$ if $a \neq \pi^*(i)$ and $d_{i,a} = 0$ if $a = \pi^*(i)$.

Note that $\sum_{a \in \mathcal{A}} \hat{\lambda}_{a,i} \geq \xi_i$ for all $i \in \mathcal{S}$. We have

$$\mathbf{P}(\hat{\pi}^{Tr} \neq \pi^*) = \mathbf{P}\left(\exists (a,i) \text{ s.t. } \pi^*(i) \neq a, \hat{\lambda}_{a,i} > \hat{\lambda}_{a',i}, \forall a' \in \mathcal{A}\right)$$

$$\leq \mathbf{P}\left(\exists (a,i) \text{ s.t. } \pi^*(i) \neq a, \hat{\lambda}_{a,i} > \frac{1}{|\mathcal{A}|}\sum_{a' \in \mathcal{A}} \hat{\lambda}_{a',i}\right)$$

$$\leq \mathbf{P}\left(\exists (a,i) \text{ s.t. } \pi^*(i) \neq a, \hat{\lambda}_{a,i} > \frac{\xi_i}{|\mathcal{A}|}\right)$$

$$\leq \mathbf{P}\left(\sum_{(a,i):\pi^*(i) \neq a} \hat{\lambda}_{a,i} d_{i,a} \geq \frac{(\min_{i \in \mathcal{S}} \xi_i)}{|\mathcal{A}|} \min_{(a,i):\pi^*(i) \neq a} d_{i,a}\right)$$

$$= \mathbf{P}\left(\sum_{(a,i)} \hat{\lambda}_{a,i} d_{i,a} \geq \frac{(\min_{i \in \mathcal{S}} \xi_i)\bar{d}}{|\mathcal{A}|}\right),$$

where the first inequality uses the positivity of $d_{i,a}$ where $\pi^*(i) \neq a$.

Let $\epsilon = \frac{(\min_{i \in \mathcal{S}} \xi_i)\bar{d}}{|\mathcal{A}|}$ and apply Theorem 3. We obtain that if $T = \Omega\left(\frac{|\mathcal{S}|^4|\mathcal{A}|^4\sigma^2}{(1-\gamma)^4\bar{d}^2} \ln\left(\frac{1}{\delta}\right)\right)$, we have

$$\mathbf{P}(\hat{\pi}^{Tr} \neq \pi^*) \leq \mathbf{P}\left(\sum_{(a,i)} \hat{\lambda}_{a,i} d_{i,a} \geq \frac{(\min_{i \in \mathcal{S}} \xi_i)\bar{d}}{|\mathcal{A}|}\right) \leq \delta,$$

for all $\delta > 0$. ∎

### 6.2 Analysis of the SPD-fMDP Algorithm 2

Our main results for the SPD-fMDP Algorithm 2 are developed through a series of lemmas. Theorem 2 suggests a critical correspondence between the optimal multiplier $\lambda^*$ and the optimal policy $\pi^*$. In particular, the basis of $\lambda^*$ can be used to yield the optimal policy $\pi^*$ as follows:

$$\pi_h^*(i) = a, \quad \text{if } \lambda_{h,a}^*(i) > 0.$$

In other words, *finding the optimal policy is equivalent to finding the basis of the optimal dual solution.* We denote by $\mathscr{F}_k$ the collection of all random variables generated up to iteration $k$.

### 6.2.1 PRELIMINARY LEMMAS

Next we provide characterization of the primal and dual solutions. These results will be used to regularize the iterates generated by our stochastic algorithms. They will also be used to establish the finite-sample error bound.

**Lemma 5** *Suppose that $(v^*, \lambda^*)$ is a pair of primal and dual solutions to the linear programs (8), (9). Then*

$$\|v_h^*\|_\infty \leq (H-h)\sigma, \quad \|v_h^*\|_2 \leq \sqrt{n}(H-h)\sigma, \quad \|\lambda_h^*\|_2 \leq \|\lambda_h^*\|_1 = \sum_{\tau=0}^{h} \|\xi_\tau\|_1.$$

*When we choose $\xi_0 = \frac{e}{H}$ and $\xi_h = \frac{e}{(H-h)(H-h+1)}$ for $h \neq 0$, we have $\|\lambda_h^*\|_2 \leq \|\lambda_h^*\|_1 = \frac{n}{H-h}$.*

*Proof.* We first prove that $\|v_h^*\|_\infty \leq (H-h)\sigma$. As shown before, the optimal value function satisfies the condition $(\mathbf{I} - \Pi_{\pi^*})v^* = R_{\pi^*}$. Solving the last $|\mathcal{S}|$ equations gives us $v_{H-1}^* = r_{H-1}^*$. Replacing it into the next last $|\mathcal{S}|$ equations, we have $\|v_{H-2}^*\|_\infty = \|P_\pi^* v_{H-1}^* + r_{H-1}^*\|_\infty \leq \|P_\pi^* v_{H-1}^*\|_\infty + \|r_{H-1}^*\|_\infty \leq 2\sigma$, where $\sigma$ is the upperbound of each component of $r$. Similarly, we have $\|v_h^*\|_\infty \leq (H-h)\sigma$ for $h \in [H]$, which is the first part of the lemma. The proof of the second part is straighforward as $\|v_h^*\|_2 \leq \sqrt{n}\|v_h^*\|_\infty \leq \sqrt{n}(H-h)\sigma$.

We then prove that $\|\lambda_h^*\|_1 = \sum_{\tau=0}^{h} \|\xi_\tau\|_1$. By Theorem 2, we have

$$(\mathbf{I} - \Pi_{\pi^*}^T)\left(\lambda_{\pi^*(i)}^*(i)\right)_{i \in \mathcal{S}_{[H]}} = \xi.$$

For the simplicity of notation, we decompose $\left(\lambda_{\pi^*(i)}^*(i)\right)_{i \in \mathcal{S}_{[H]}}$ into $(\lambda_0^{*T}, \lambda_1^{*T}, \ldots, \lambda_H^{*T})^T$ where $\lambda_0^*$ are the first $\mathcal{S}$ elements of $\left(\lambda_{\pi^*(i)}^*(i)\right)_{i \in \mathcal{S}_{[H]}}$ and so forth. Solving the first $|\mathcal{S}|$ equations, we have $\lambda_0^* = \xi_0$. Replacing it into the next $|\mathcal{S}|$ equations, we get

$$\|\lambda_1^*\|_1 = e^T \lambda_1^* = e^T(\xi_1 + P_{\pi^*}^T \lambda_0^*) = e^T \xi_1 + e^T P_{\pi^*}^T \lambda_0^*$$
$$= \|\xi_1\|_1 + e^T \lambda_0^* = \|\xi_0\|_1 + \|\xi_1\|_1.$$

By induction, we have $\|\lambda_h^*\|_1 = \sum_{\tau=0}^{h} \|\xi_\tau\|_1$. Therefore, $\|\lambda_h^*\|_2 \leq \|\lambda_h^*\|_1 = \sum_{\tau=0}^{h} \|\xi_\tau\|_1$. When we choose $\xi_0 = \frac{e}{H}$ and $\xi_h = \frac{e}{(H-h)(H-h+1)}$ for $h \neq 0$, we have $\sum_{\tau=0}^{h} \|\xi_\tau\|_1 = \frac{n}{H} + \frac{n}{H(H-1)} + \cdots = \frac{n}{H-h}$. To this point, we proved Lemma 2. ∎

In order to prove Theorem 6, we also need a lowerbound on the value of the dual variables.

**Lemma 6** *Suppose that $(v^*, \lambda^*)$ is a pair of primal and dual solutions to the linear programs (8), (9). Then*

$$\sum_{a \in \mathcal{A}} \lambda_{h,a}^*(i) \geq \xi_h(i), \quad \forall\, h \in [H], i \in \mathcal{S}_h.$$

*Proof.* We observe that $\sum_{a \in \mathcal{A}} \lambda^*_{h,a}(i) = \lambda^*_{\pi^*(i)}(i), \forall\ h \in [H], i \in \mathcal{S}_h$. So to prove the lemma, we only need to show $\lambda^*_{\pi^*(i)} \geq \xi_h(i)$. From the notation that $\left(\lambda^*_{\pi^*(i)}(i)\right)_{i \in \mathcal{S}_{[H]}} = (\lambda_0^{*T}, \lambda_1^{*T}, \ldots, \lambda_H^{*T})^T$, it is equivalent to show that $\lambda^*_h \geq \xi_h$. Using the same formulation as the proof Lemma 2, we have $\lambda_0^* = \xi_0$. Replacing it into the next $n$ equations, we get

$$\lambda_1^* = \xi_1 + P_{\pi_h^*} \lambda_0^* \geq \xi_1, \quad \forall\ i \in \mathcal{S}_1$$

where $P_{\pi_h^*}$ is the transition probability at stage $h$ under policy $\pi$ and the inequality is due to that $\lambda_0^* \geq 0$. Then by induction, we have $\sum_{a \in \mathcal{A}} \lambda^*_{h,a}(i) \geq \xi_h(i), \forall\ h \in [H], i \in \mathcal{S}_h$. ∎

The following is the inequality used to bound the infinity norm of the error difference.

**Lemma 7** *We have the following inequality,*

$$\|(\mathbf{I} - \hat{\Pi})v\|_1 \geq \|v\|_\infty.$$

*Proof.* Let $v = (v_0^T, \ldots, v_{H-1}^T)^T$ and $w = (\mathbf{I} - \hat{\Pi})v$. Solving the last $n$ equations give us $w_{H-1} = v_{H-1}$, where $w_{H-1}$ is the last $|\mathcal{S}|$ components of $w$. Replacing the results into the next last $n$ equations, we get $v_{H-2} = w_{H-2} + Pv_{H-1}$ where $P$ is a stochastic matrix. As a result, we have $\|v_{H-2}\|_\infty \leq \|w_{H-2}\|_\infty + \|P\|_\infty \|v_{H-1}\|_\infty \leq \|w_{H-2}\|_1 + \|w_{H-1}\|_1$. Similarly, we have $\|v_h\|_\infty \leq \sum_{i=h}^{H-1} \|w_i\|_1$. Therefore, we have $\|v\|_\infty \leq \|w\|_1 = \|(\mathbf{I} - \hat{\Pi})v\|_1$. ∎

**Lemma 8** *We denote for short that*

$$\mathcal{E}_k = \sum_{h=0}^{H-1} \frac{n}{(H-h)^2 \sigma} \|v_h^k - v_h^*\|_2^2 + \sum_{h=0}^{H-1} (H-h)^2 m\sigma \sum_{a \in \mathcal{A}} \|\lambda_{h,a}^k - \lambda_{h,a}^*\|_2^2$$

*and*

$$\mathcal{G}_k = \sum_{h=0}^{H-1} \sum_{a \in \mathcal{A}} (\lambda_{h,a}^k)^T (r_a + P_a v_{h+1}^* - v_h^*).$$

*Then the SPD-fMDP Algorithm 2 satisfies for all $k$ with probability 1 that*

$$\mathcal{E}_k \leq \mathcal{O}(mn^2 H \sigma), \quad \mathbf{E}\left[\mathcal{E}_{k+1} \mid \mathscr{F}_k\right] \leq \mathcal{E}_k - \frac{2}{\sqrt{k+1}} \mathcal{G}_k + \frac{1}{k+1} \mathcal{O}\left(mn^2 H \sigma\right). \tag{18}$$

*Proof.* We first show that the duality gap comes naturally as we analyze the distance between the optimal primal-dual variables and the iterates outputed by the SPD-fMDP Algorithm 2. The SPD-fMDP Algorithm 2 updates the iterate according to the saddle point problem. However, the algorithm does not have access to the transition probability and the expected rewards. Instead, it will estimate the probability and reward from the sampling oracle in each iteration.

Let $(i_k, j_k, a_k, \hat{r}_{i_k j_k a_k})$ be the sample given by the oracle at iteration $k$. Let $\hat{P}_a^{(k)}$ be $n \times n$ random matrices indexed by action $a$ with $\hat{P}_{a_k}^{(k)}(i_k, j_k) = mn$ and 0 in all the other entries.

Let $\hat{\mathbf{I}}_a^{(k)}$ be $n \times n$ random matrices with $\hat{\mathbf{I}}_{a_k}^{(k)}(i_k) = mn$ and $\hat{\mathbf{I}}_a^{(k)}(i) = 0$ for $a \neq a_k$ or $i \neq i_k$ where $\mathbf{I}$ is the identity matrix. Define $\hat{\Gamma}_a^{(k)}$ to be

$$\hat{\Gamma}_a^{(k)} = \begin{bmatrix} \hat{\mathbf{I}}_a & -\hat{P}_a & 0 & \ldots & 0 \\ 0 & \hat{\mathbf{I}}_a & -\hat{P}_a & \ldots & 0 \\ \vdots & \vdots & \vdots & \ddots & \vdots \\ 0 & 0 & 0 & \ldots & -\hat{P}_a \\ 0 & 0 & 0 & \ldots & \hat{\mathbf{I}}_a \end{bmatrix},$$

with dimension $mn \times mn$. By our construction, we have $\mathbf{E}[\hat{\Gamma}_a^{(k)}] = \mathbf{I} - \Pi_a$ for all action $a$. With some abuse of notation, we also use $\hat{\mathbf{I}}_a^{(k)}$ to denote the diagonal of $\hat{\Gamma}_a^{(k)}$ such that $\hat{\Gamma}_a^{(k)} = \hat{\mathbf{I}}_a^{(k)} - \hat{\Pi}_a$. Define $\hat{r}_a^{(k)}$ to be the vectors with $\hat{r}_{a_k}^{(k)}(i_k) = mn\hat{r}_{i_k j_k a_k}$ and 0 in all the other entries so that $\mathbf{E}[\hat{r}_a^{(k)}] = r$. Let $\hat{R}_a^{(k)} = (\hat{r}_a^{(k)}, \ldots, \hat{r}_a^{(k)})$ be $H$ replicas of $\hat{r}_a^{(k)}$ and $R_a = (r_a, \ldots, r_a)$. Let $\hat{\xi}_0(i_k) = \frac{mn}{H}$ and $\hat{\xi}_0(i) = 0 \ \forall \ i \neq i_k$. Let $\hat{\xi}_h(i_k) = \frac{mn}{(H-h)(H-h+1)}$ and $\hat{\xi}_h(i_k) = 0 \ \forall \ i \neq i_k$. We use the notation $\hat{\xi}$ to denote $(\hat{\xi}_0^T, \hat{\xi}_1^T, \ldots, \hat{\xi}_{H-1}^T)^T$. In the following, we will drop the superscript $(k)$ if there is no ambiguity.

From the above notation, the update step of the SPD-fMDP Algorithm 2 is equivalent to

$$v_h^{k+1} = \Pi_\mathcal{V} \left( v_h^k - (H-h)^2 \frac{\sigma \gamma_{k+1}}{n} \left( \hat{\xi} + \sum_{a \in \mathcal{A}} \left( \hat{P}_a^T \lambda_{h-1,a}^k - \hat{\mathbf{I}}_a \lambda_{h,a}^k \right) \right) \right)$$

$$\lambda_{h,a}^{k+1} = \Pi_{\Xi \cap \Delta} \left( \lambda_{h,a}^k + \frac{1}{(H-h)^2} \frac{\gamma_{k+1}}{m\sigma} (\hat{P}_a v_{h+1}^k - \hat{\mathbf{I}}_a v_h^k + \hat{r}_a) \right) \forall a \in \mathcal{A},$$

where $\gamma_k = \frac{1}{\sqrt{k}}$.

Analyzing the distance between the iterates and the optimal solution, we have

$$\sum_{h=0}^{H-1} \frac{n}{(H-h)^2 \sigma} \|v_h^{k+1} - v_h^*\|_2^2 + \sum_{h=0}^{H-1} (H-h)^2 m\sigma \sum_{a \in \mathcal{A}} \|\lambda_{h,a}^{k+1} - \lambda_{h,a}^*\|_2^2$$

$$\leq \sum_{h=0}^{H-1} \frac{n}{(H-h)^2 \sigma} \left\| v_h^k - v_h^* - (H-h)^2 \frac{\sigma \gamma_{k+1}}{n} \left( \hat{\xi} + \sum_{a \in \mathcal{A}} \left( \hat{P}_a^T \lambda_{h-1,a}^k - \hat{\mathbf{I}}_a \lambda_{h,a}^k \right) \right) \right\|_2^2$$

$$+ \sum_{h=0}^{H-1} (H-h)^2 m\sigma \sum_{a \in \mathcal{A}} \left\| \lambda_{h,a}^k - \lambda_{h,a}^* + \frac{\gamma_{k+1}}{(H-h)^2 m\sigma} (\hat{P}_a v_{h+1}^k - \hat{\mathbf{I}}_a v_h^k + \hat{r}_a) \right\|_2^2$$

$$= \sum_{h=0}^{H-1} \left( \frac{n}{(H-h)^2 \sigma} \|v_h^k - v_h^*\|_2^2 + (H-h)^2 m\sigma \sum_{a \in \mathcal{A}} \|\lambda_a^k - \lambda_a^*\|_2^2 \right) - 2\gamma_{k+1} (\hat{\xi} - \sum_{a \in \mathcal{A}} \hat{\Gamma}_a^T \lambda_a^k)^T (v^k - v^*)$$

$$+ 2\gamma_{k+1} \sum_{a \in \mathcal{A}} (\hat{R}_a - \hat{\Gamma}_a v^k)^T (\lambda_a^k - \lambda_a^*) + \frac{\sigma \gamma_{k+1}^2}{n} \Phi_1^k + \frac{\gamma_{k+1}^2}{m\sigma} \sum_{a \in \mathcal{A}} \Phi_a^k,$$

(19)

where $\Phi_1^k = \sum_{h=0}^{H-1}(H-h)^2\|\hat{\xi} + \sum_{a\in\mathcal{A}}(\hat{P}_a^T\lambda_{h-1,a}^k - \hat{\mathbf{I}}_a\lambda_{h,a}^k)\|_2^2$, $\Phi_a^k = \sum_{h=0}^{H-1}\frac{1}{(H-h)^2}\|\hat{P}_av_{h+1}^k - \hat{\mathbf{I}}_av_h^k + \hat{r}_a\|_2^2$ and the first inequality is due to the property of the projection. In what follows, we analyze the righthandside of the preceding inequality.

We first analyze the second term and the third term of the above inequality. Observe that

$$(\hat{\xi} - \sum_{a\in\mathcal{A}}\hat{\Gamma}_a^T\lambda_a^k)^T(v^k - v^*) - \sum_{a\in\mathcal{A}}(\hat{R}_a - \hat{\Gamma}_av^k)^T(\lambda_a^k - \lambda_a^*)$$
$$= \hat{\xi}(v^k - v^*) + \sum_{a\in\mathcal{A}}\left(\lambda_a^{*T}(\hat{R}_a - \hat{\Gamma}_av^k) - (\lambda_a^k)^T(\hat{R}_a - \hat{\Gamma}_av^*)\right). \tag{20}$$

For any $v$, let $f(v) = \xi^T v$. Then $\mathbf{E}\left[\hat{\xi}(v^k - v^*)\right] = f(v^k) - f(v^*)$. Taking conditional expectation on (20), we have

$$\mathbf{E}\left[\sum_{a\in\mathcal{A}}\lambda_a^{*T}(\hat{R}_a - \hat{\Gamma}_a x^k)\bigg|\mathscr{F}_k\right] = \sum_{a\in\mathcal{A}}\lambda_a^{*T}(R_a - (\mathbf{I} - \Pi_a)v^k)$$
$$= \sum_{a\in\mathcal{A}}\lambda_a^{*T}R_a - \xi^Tv^k = -f(v^k) + f(v^*),$$

where the first equality is due to the definition of $\hat{\Gamma}_a, \hat{R}_a$ and the second equality uses the dual feasibility that $\sum_{a\in\mathcal{A}}(\mathbf{I} - \Pi_a^T)\lambda_a^* = \xi$. Also by taking conditional expectation, we have

$$\mathbf{E}\left[\sum_{a\in\mathcal{A}}-(\lambda_a^k)^T(\hat{R}_a - \hat{\Gamma}_a x^*)\bigg|\mathscr{F}_k\right] = \sum_{a\in\mathcal{A}}(\lambda_a^k)^T(v^* - \Pi_av^* - R_a).$$

Let $\mathcal{E}_k = \sum_{h=0}^{H-1}\frac{n}{(H-h)^2\sigma}\|v_h^k - v_h^*\|_2^2 + \sum_{h=0}^{H-1}(H-h)^2m\sigma\sum_{a\in\mathcal{A}}\|\lambda_{h,a}^k - \lambda_{h,a}^*\|_2^2$. Taking conditional expectation on (19) and substituting all the above equations, we have

$$\mathbf{E}[\mathcal{E}_{k+1}|\mathscr{F}_k] \leq \mathcal{E}_k - 2\gamma_{k+1}\sum_{a\in\mathcal{A}}(\lambda_a^k)^T(v^* - \Pi_av^* - R_a)$$
$$+ \frac{\sigma\gamma_{k+1}^2}{n}\mathbf{E}[\Phi_1^k|\mathscr{F}_k] + \frac{\gamma_{k+1}^2}{m\sigma}\mathbf{E}[\sum_{a\in\mathcal{A}}\Phi_a^k|\mathscr{F}_k]. \tag{21}$$

Next, we analyze $\mathbf{E}[\Phi_1^k|\mathscr{F}_k]$. Note that

$$\mathbf{E}[\Phi_1^k|\mathscr{F}_k] = \sum_{h=0}^{H-1}(H-h)^2\|\hat{\xi}_h + \sum_{a\in\mathcal{A}}(\hat{P}_a^T\lambda_{h-1,a}^k - \hat{\mathbf{I}}_a\lambda_{h,a}^k)\|_2^2$$
$$\leq 2H + \sum_{h=0}^{H-1}(H-h)^2\left(4\mathbf{E}\left[\sum_{a\in\mathcal{A}}\hat{\mathbf{I}}_a\lambda_{h,a}^k\bigg|\mathscr{F}_k\right] + 4\mathbf{E}\left[\left\|\sum_{a\in\mathcal{A}}\hat{P}_a^T\lambda_{h-1,a}^k\right\|^2\bigg|\mathscr{F}_k\right]\right), \tag{22}$$

where the last inequality is due to Lemma 5 and the basic inequality $(x+y)^2 \leq 2x^2 + 2y^2$.
To bound the second term of (22), we observe by Lemma 5 that

$$\sum_{h=0}^{H-1}(H-h)^2 \mathbf{E}\left[\left\|\sum_{a\in\mathcal{A}}\hat{\mathbf{I}}_a\lambda_{h,a}^k\right\|_2^2\Big|\mathcal{F}_k\right] = \sum_{h=0}^{H-1}(H-h)^2 \mathbf{E}_{a_{k+1}}\mathbf{E}_{i_{k+1}}\mathbf{E}\left[m^2n^2\lambda_{h,a_{k+1}}^k(i_{k+1})^2|\mathcal{F}_k, a_{k+1}, i_{k+1}\right]$$

$$= \sum_{h=0}^{H-1}(H-h)^2 \sum_{a=a_1}^{a_m}\sum_{i=1}^{n}\frac{1}{mn}\cdot m^2n^2\lambda_{h,a}^k(i)^2$$

$$\leq \sum_{h=0}^{H-1}(H-h)^2 mn\|\lambda_h\|_1^2 \leq mn^3H.$$

We apply a similar analysis to the third term of (22) and get

$$\sum_{h=0}^{H-1}(H-h)^2 \mathbf{E}\left[\left\|\sum_{a\in\mathcal{A}}\hat{P}_a^T\lambda_{h-1,a}^k\right\|^2\Big|\mathcal{F}_k\right]$$

$$= \sum_{h=0}^{H-1}(H-h)^2 \mathbf{E}_{a_{k+1}}\mathbf{E}_{i_{k+1}}\mathbf{E}\left[\|\sum_{a\in\mathcal{A}}\hat{P}_a^T\lambda_{h-1,a}^k\|^2|\mathcal{F}_k, a_{k+1}, i_{k+1}\right]$$

$$= \sum_{h=0}^{H-1}(H-h)^2 \mathbf{E}_{a_{k+1}}\mathbf{E}_{i_{k+1}}\mathbf{E}\left[m^2n^2\lambda_{h-1,a_{k+1}}^k(i_{k+1})^2|\mathcal{F}_k, a_{k+1}, i_{k+1}\right].$$

Appying Lemma 5 gives us $\sum_{h=0}^{H-1}(H-h)^2\mathbf{E}\left[\left\|\sum_{a\in\mathcal{A}}\hat{P}_a^T\lambda_{h-1,a}^k\right\|^2|\mathcal{F}_k\right] \leq mn^3H$.

Plugging the above bound to (22), we have

$$\mathbf{E}[\Phi_1^k|\mathcal{F}_k] = \mathcal{O}(mn^3H). \tag{23}$$

Now we bound $\mathbf{E}[\sum_{a\in\mathcal{A}}\Phi_a^k|\mathcal{F}_k]$. Observe that

$$\mathbf{E}\left[\sum_{a\in\mathcal{A}}\Phi_a^k|\mathcal{F}_k\right] = \mathbf{E}\left[\sum_{a\in\mathcal{A}}\sum_{h=0}^{H-1}\frac{1}{(H-h)^2}\|\hat{P}_av_{h+1}^k - \hat{\mathbf{I}}_av_h^k + \hat{r}_a\|_2^2\right]$$

$$\leq \sum_{h=0}^{H-1}\frac{1}{(H-h)^2}\left(2\mathbf{E}\left[\sum_{a\in\mathcal{A}}\|\hat{P}_av_{h+1}^k - \hat{\mathbf{I}}_av_h^k\|^2|\mathcal{F}_k\right] + 2\mathbf{E}\left[\sum_{a\in\mathcal{A}}\|\hat{r}_a\|^2|\mathcal{F}_k\right]\right). \tag{24}$$

By Lemma 5, we can bound the first term of (24) as

$$2\mathbf{E}\left[\sum_{a\in\mathcal{A}}\|\hat{P}_av_{h+1}^k - \hat{\mathbf{I}}_av_h^k\|^2|\mathcal{F}_k\right] \leq 4\mathbf{E}\left[\sum_{a\in\mathcal{A}}\|\hat{\mathbf{I}}_av_h^k\|^2|\mathcal{F}_k\right] + 4\mathbf{E}\left[\sum_{a\in\mathcal{A}}\|\hat{P}_av_{h+1}^k\|^2|\mathcal{F}_k\right]$$

$$\leq 4n^2m^2\|v_h^k\|_\infty^2 + 4\sum_{a_{k+1}\in\mathcal{A}}\sum_{i_{k+1}\in\mathcal{S}}\frac{1}{mn}\cdot m^2n^2\|v_{h+1}^k\|_\infty^2$$

$$\leq 8n^2m^2\|v_h^k\|_\infty^2 \leq 8m^2n^2(H-h)^2\sigma^2.$$

By the definition of $\sigma$, we can bound the second term as

$$\mathbf{E}[\sum_{a\in\mathcal{A}}\|\hat{r}_a\|^2|\mathscr{F}_k] = \mathbf{E}_{a_{k+1}}\mathbf{E}_{i_{k+1}}\mathbf{E}[\sum_{a\in\mathcal{A}}\|\hat{r}_a\|^2|\mathscr{F}_k, a_{k+1}, i_{k+1}]$$
$$\leq \sum_{a_{k+1}\in\mathcal{A}}\sum_{i_{k+1}\in\mathcal{S}} mn\sigma^2 \leq m^2n^2\sigma^2.$$

Applying the preceding results to (24), we have

$$\mathbf{E}\left[\sum_{a\in\mathcal{A}}\Phi_a^k|\mathscr{F}_k\right] \leq \sum_{h=0}^{H-1}\frac{1}{(H-h)^2}(8m^2n^2(H-h)^2\sigma^2 + m^2n^2\sigma^2) = \mathcal{O}(m^2n^2H\sigma^2). \quad (25)$$

Let $\mathcal{G}_k = \sum_{a\in\mathcal{A}}(\lambda_a^k)^T(v^* - \Pi_a v^* - R_a)$. We apply equations (22), (24) and $\gamma_k = \frac{1}{\sqrt{k}}$ to (21), which gives

$$\mathbf{E}[\mathcal{E}_{k+1}|\mathcal{F}_k] \leq \mathcal{E}_k - \frac{2}{\sqrt{k+1}}\mathcal{G}_k + \frac{1}{k+1}\mathcal{O}\left(\frac{\sigma}{n}\cdot mn^3H + \frac{1}{m\sigma}\cdot m^2n^2H\sigma^2\right), \quad (26)$$

which completes the proof of the first claim.

Note that in each iteration, we project the iterate onto $\mathcal{V}$ and $\Xi\cap\Lambda$. As a result, we can apply Lemma 5 and get

$$\mathcal{E}_k = \sum_{h=0}^{H-1}\frac{n}{(H-h)^2\sigma}\|v_h^k - v_h^*\|_2^2 + \sum_{h=0}^{H-1}(H-h)^2m\sigma\sum_{a\in\mathcal{A}}\|\lambda_{h,a}^k - \lambda_{h,a}^*\|_2^2$$
$$\leq \sum_{h=0}^{H-1}\frac{n}{(H-h)^2\sigma}\cdot 4n\sigma^2(H-h)^2 + \sum_{h=0}^{H-1}(H-h)^2m\sigma\cdot\frac{4n^2}{(H-h)^2} = \mathcal{O}\left(mn^2H\sigma\right),$$

for all $k$, which completes the whole claim. $\blacksquare$

**Lemma 9** *We have*

$$\frac{\sqrt{k+1}}{2}\left(\mathbf{E}\left[(\mathcal{E}_{k+1} - \mathbf{E}\left[\mathcal{E}_{k+1}\mid\mathscr{F}_k\right])^2\right]\right)^{1/2} \leq \mathcal{O}(n^2mH),$$

*and there exists $M > 0$ such that*

$$\frac{\sqrt{k+1}}{2}(\mathcal{E}_{k+1} - \mathbf{E}\left[\mathcal{E}_{k+1}\mid\mathscr{F}_k\right]) \leq M, \qquad w.p.1.$$

*Proof.* To see this, we derive

$$\mathcal{E}_{k+1} - \mathcal{E}_k = \sum_{h=0}^{H-1}\frac{n}{(H-h)^2\sigma}\|v_h^{k+1} - v_h^*\|_2^2 + \sum_{h=0}^{H-1}(H-h)^2m\sigma\sum_{a\in\mathcal{A}}\|\lambda_{h,a}^{k+1} - \lambda_{h,a}^*\|_2^2$$
$$- \sum_{h=0}^{H-1}\frac{n}{(H-h)^2\sigma}\|v_h^k - v_h^*\|_2^2 + \sum_{h=0}^{H-1}(H-h)^2m\sigma\sum_{a\in\mathcal{A}}\|\lambda_{h,a}^k - \lambda_{h,a}^*\|_2^2.$$

Observe that
$$\|v_h^{k+1} - v_h^*\|_2^2 - \|v_h^k - v_h^*\|_2^2 = 2(v_h^{k+1} - v_h^k)^T(v_h^k - v_h^*) + \|v_h^{k+1} - v_h^k\|_2^2$$
$$\leq 4D_{h,v}\|v_h^{k+1} - v_h^k\|_2 + \|v_h^{k+1} - v_h^k\|_2^2,$$

where we denote for short that $D_{h,v}^2 = n(H-h)^2\sigma^2$ and the first inequality is due to the observation that $(v_h^{k+1} - v_h^k)^T(v_h^k - v_h^*) \leq \|v_h^{k+1} - v_h^k\| \|v_h^k - v_h^*\|$. Note that by the nonexpansiveness of projection, we have

$$\mathbf{E}\left[\|v_h^{k+1} - v_h^k\|_2^2 \mid \mathscr{F}_k\right] \leq \mathbf{E}\left[(H-h)^4 \frac{\sigma^2\gamma_{k+1}^2}{n^2} \left\|\hat{\xi} + \sum_{a\in\mathcal{A}}\left(\hat{P}_a^T\lambda_{h-1,a}^k - \hat{\mathbf{I}}_a\lambda_{h,a}^k\right)\right\|_2^2 \Bigg| \mathscr{F}_k\right],$$

which is smaller than $\gamma_{k+1}^2(H-h)^2\sigma^2 mn$ by Lemma 5 and the analysis of $\mathbf{E}[\Phi_1^k|\mathscr{F}_k]$ in the previous proof.

Then

$$\mathbf{E}\left[\left(\sum_{h=0}^{H-1} \frac{n}{(H-h)^2\sigma}\left(\|v_h^{k+1} - v_h^*\|_2^2 - \|v_h^k - v_h^*\|_2^2\right)\right)^2 \Bigg| \mathscr{F}_k\right]$$
$$\leq \mathbf{E}\left[H \sum_{h=0}^{H-1} \frac{n^2}{(H-h)^4\sigma^2}\left(32D_{h,v}^2\|v_h^{k+1} - v_h^k\|_2^2 + 2\|v_h^{k+1} - v_h^k\|_2^4\right) \Bigg| \mathscr{F}_k\right]$$
$$= \mathcal{O}\left(\gamma_{k+1}^2 n^4 m H^2 \sigma^2 + \gamma_{k+1}^4\right) = \mathcal{O}\left(\gamma_{k+1}^2 n^4 m H^2 \sigma^2\right).$$

We can bound $\lambda_{h,a}^k$ in a similar way. Denote $D_{h,\lambda} = \frac{n}{H-h}$. We have

$$\mathbf{E}\left[\left(\sum_{h=0}^{H-1}(H-h)^2 m\sigma \sum_{a\in\mathcal{A}} \|\lambda_{h,a}^{k+1} - \lambda_{h,a}^*\|_2^2\right)^2 \Bigg| \mathscr{F}_k\right]$$
$$\leq \mathbf{E}\left[H \sum_{h=0}^{H-1}(H-h)^4 m^2\sigma^2\left(32D_{h,\lambda}^2\|\lambda_h^{k+1} - \lambda_h^k\|_2^2 + 2\|\lambda_h^{k+1} - \lambda_h^k\|_2^4\right) \Bigg| \mathscr{F}_k\right]$$
$$= \mathcal{O}\left(\gamma_{k+1}^2 n^4 m^2 H^2 \sigma^2 + \gamma_{k+1}^4\right) = \mathcal{O}\left(\gamma_{k+1}^2 n^4 m^2 H^2 \sigma^2\right).$$

Substituting the preceding inequalities, we have
$$\mathbf{E}\left[|\mathcal{E}_{k+1} - \mathcal{E}_k|^2 \mid \mathscr{F}_k\right] = \mathcal{O}(\gamma_{k+1}^2 n^4 m^2 H^2 \sigma^2).$$

Note that $\mathcal{E}_k \in \mathscr{F}_k$, therefore $\mathbf{E}\left[\mathcal{E}_k \mid \mathscr{F}_k\right] = \mathcal{E}_k$ and
$$\mathcal{E}_{k+1} - \mathbf{E}\left[\mathcal{E}_{k+1} \mid \mathscr{F}_k\right] = \mathcal{E}_{k+1} - \mathcal{E}_k - \mathbf{E}\left[\mathcal{E}_{k+1} - \mathcal{E}_k \mid \mathscr{F}_k\right].$$

Then we have
$$\mathbf{E}\left[|\mathcal{E}_{k+1} - \mathbf{E}\left[\mathcal{E}_{k+1} \mid \mathscr{F}_k\right]|^2\right] = \mathbf{E}\left[|\mathcal{E}_{k+1} - \mathcal{E}_k - \mathbf{E}\left[\mathcal{E}_{k+1} - \mathcal{E}_k \mid \mathscr{F}_k\right]|^2\right]$$
$$= \mathbf{E}\left[\mathbf{E}\left[|\mathcal{E}_{k+1} - \mathcal{E}_k - \mathbf{E}\left[\mathcal{E}_{k+1} - \mathcal{E}_k \mid \mathscr{F}_k\right]|^2 \mid \mathscr{F}_k\right]\right]$$
$$\leq \mathbf{E}\left[\mathbf{E}\left[|\mathcal{E}_{k+1} - \mathcal{E}_k|^2 \mid \mathscr{F}_k\right]\right]$$
$$\leq \mathcal{O}(\gamma_{k+1}^2 n^4 m^2 H^2 \sigma^2),$$

where the inequality uses the fact that the variance of a random variable is bounded by its second moment. In the second equality, the first expectation is taken over $\mathscr{F}_k$ and the second expectation is taken over all the randomness at round $k+1$. The proof of the second part is similar to the proof of Lemma 4. Denote $\Delta v_h^k$ to be $\Delta v_h^k = \hat{\xi} + \sum_{a\in\mathcal{A}} \left(\hat{P}_a^T \lambda_{h-1,a}^k - \hat{\mathbf{I}}_a \lambda_{h,a}^k\right)$. We can show that

$$\sum_{h=0}^{H-1} \frac{n}{(H-h)^2\sigma}\|v_h^{k+1} - v_h^*\|_2^2 - \mathbf{E}\left[\sum_{h=0}^{H-1} \frac{n}{(H-h)^2\sigma}\|v_h^{k+1} - v_h^*\|_2^2\right]$$

$$\leq \sum_{h=0}^{H-1} \frac{n}{(H-h)^2\sigma}\|v_h^k - v_h^* - \frac{(H-h)^2\sigma}{\sqrt{k}n}\Delta v_h^k\|_2^2 - \mathbf{E}\left[\sum_{h=0}^{H-1} \frac{n}{(H-h)^2\sigma}\|v_h^k - v_h^* - \frac{(H-h)^2\sigma}{\sqrt{k}n}\Delta v_h^k\|_2^2\right]$$

$$\leq \mathcal{O}\left(\frac{1}{\sqrt{k}}\sum_{h=0}^{H-1}\|v_h^k - v_h^*\|_2\|\Delta v_h^k\|_2\right) \leq \mathcal{O}\left(\frac{\sigma m n^{5/2}H}{\sqrt{k}}\right).$$

Similarly, we can show that

$$\sum_{h=0}^{H-1}(H-h)^2 m\sigma \left(\sum_{a\in\mathcal{A}}\|\lambda_{h,a}^{k+1} - \lambda_{h,a}^*\|_2^2 - \mathbf{E}\left[\sum_{a\in\mathcal{A}}\|\lambda_{h,a}^{k+1} - \lambda_{h,a}^*\|_2^2\right]\right) \leq \mathcal{O}\left(\frac{\sigma m n^2 H}{\sqrt{k}}\right).$$

Let $M = \frac{C\sigma m n^{5/2} H}{\sqrt{k}}$ for some constant $C$ and we can see that the difference is bounded by the constant $M$. ∎

6.2.2 PROOF OF THEOREM 6, 7 AND 8

**Proof of Theorem 6** Rearranging the terms of (18), we have

$$\mathcal{G}_k \leq \frac{\sqrt{k+1}}{2}(\mathcal{E}_k - \mathbf{E}\left[\mathcal{E}_{k+1} \mid \mathscr{F}_k\right]) + \frac{1}{2\sqrt{k+1}}\mathcal{O}\left(mn^2 H\sigma\right).$$

Summing over $k$ and taking average, we have

$$\frac{1}{T}\sum_{k=1}^{T}\mathcal{G}_k \leq \frac{1}{T}\sum_{k=1}^{T}\frac{\sqrt{k+1}}{2}(\mathcal{E}_k - \mathbf{E}\left[\mathcal{E}_{k+1} \mid \mathscr{F}_k\right]) + \frac{1}{2T}\sum_{k=1}^{T}\frac{1}{\sqrt{k+1}}\mathcal{O}\left(mn^2 H\sigma\right)$$

$$\leq \frac{1}{T}\sum_{k=1}^{T}\frac{\sqrt{k+1}-\sqrt{k}}{2}\mathcal{E}_k + \frac{1}{T}\sum_{k=1}^{T}\frac{\sqrt{k+1}}{2}(\mathcal{E}_{k+1} - \mathbf{E}\left[\mathcal{E}_{k+1} \mid \mathscr{F}_k\right]) + \frac{\mathcal{E}_1}{2T} + \mathcal{O}\left(\frac{mn^2 H\sigma}{\sqrt{T}}\right).$$

Let us construct a sequence of random variables $\{M_t\}$ given by

$$M_{t+1} = \sum_{k=1}^{t}\frac{\sqrt{k+1}}{2}\left(\mathcal{E}_{k+1} - \mathbf{E}\left[\mathcal{E}_{k+1} \mid \mathscr{F}_k\right]\right).$$

By the construction of $M_t$, we have $\mathbf{E}\left[M_{t+1} \mid \mathscr{F}_t\right] = M_t$, which implies that $M_t$ is a martingale. According to Lemma 9, we also have $\frac{\sqrt{k+1}}{2}(\mathcal{E}_{k+1} - \mathbf{E}\left[\mathcal{E}_{k+1} \mid \mathscr{F}_k\right]) \leq M$ and $\frac{\sqrt{k+1}}{2}\left(\mathbf{E}\left[(\mathcal{E}_{k+1} - \mathbf{E}\left[\mathcal{E}_{k+1} \mid \mathscr{F}_k\right])^2\right]\right)^{1/2} \le \mathcal{O}(n^2 m H \sigma)$ with probability 1. As a result, $M_t$ is a martingale with bounded difference and its difference has bounded second moment. We apply the Bernstein inequality and obtain for any $\epsilon > 0$ that

$$\mathbf{P}\left(\frac{1}{T}M_T \ge \epsilon\right) \le \exp\left(-\frac{T\epsilon^2}{2n^4 m^2 H^2 \sigma^2 + (2/3)M\epsilon}\right).$$

By taking $T \ge 2\max\left\{\frac{2n^4 m^2 H^2 \sigma^2 \ln(1/\delta)}{\epsilon^2}, \frac{2M\ln(1/\delta)}{3\epsilon}\right\}$, we obtain $\mathbf{P}\left(\frac{1}{T}\sum_{k=1}^T M_k \ge \epsilon\right) \le \delta$. By Lemma 8, we know that $\mathcal{E}_{k+1}$ is uniformly bounded by $\mathcal{O}(mn^2 H \sigma)$. So with probability at least $1 - \delta$, we have

$$\frac{1}{T}\sum_{k=1}^T \mathcal{G}_k \le \frac{1}{2\sqrt{T}}\mathcal{O}(n^2 m H \sigma) + \mathcal{O}(\epsilon) + \mathcal{O}\left(\frac{mn^2 H \sigma}{\sqrt{T}}\right).$$

Therefore by taking $\xi_0 = \frac{e}{H}$ and $\xi_h = \frac{e}{(H-h)(H-h+1)}$ for $h \ne 0$, we obtain

$$\frac{1}{T}\sum_{k=1}^T \mathcal{G}_k = \mathcal{O}\left(\frac{n^2 m H \sigma}{\sqrt{T}}\right) + \mathcal{O}(\epsilon),$$

with probability at least $1 - \delta$.

When $\epsilon$ is smaller than $\frac{3n^4 m^2 H^2 \sigma^2}{M}$, we have $T \ge \frac{4n^4 m^2 H^2 \sigma^2 \ln(1/\delta)}{\epsilon^2}$. Plugging the relationship into the preceding equality, we have

$$\frac{1}{T}\sum_{k=1}^T \mathcal{G}_k = \mathcal{O}(\epsilon),$$

with probability at least $1 - \delta$. ∎

**Proof of Theorem 7** In the following, we show that the duality gap for the bilinear saddle point problem gives a bound on the efficiency loss of the randomized policy $\hat{\pi}$. Note that $\hat{\lambda} = \frac{1}{T}\sum_{k=1}^T \lambda^k$. So we have

$$\left(\frac{1}{T}\sum_{k=1}^T \mathcal{G}_k\right) = \sum_{h=0}^{H-1}\sum_{a \in \mathcal{A}} (\hat{\lambda}_{h,a})^T (r_a + P_a v^*_{h+1} - v^*_h)$$

$$= \sum_{h=0}^{H-1}\sum_{a \in \mathcal{A}, i \in \mathcal{S}} \hat{\lambda}_{h,a}(i)(r_a(i) + P_{a,i} v^*_{h+1} - v^*_h(i))$$

$$= \sum_{h=0}^{H-1}\sum_{i \in \mathcal{S}} \left(\sum_{a' \in \mathcal{A}} \hat{\lambda}_{h,a'}(i)\right) \sum_{a \in \mathcal{A}} \frac{\hat{\lambda}_{h,a}(i)}{\sum_{a' \in \mathcal{A}} \hat{\lambda}_{h,a'}(i)}(r_a(i) + P_{a,i} v^*_{h+1} - v^*_h(i)).$$

We denote

$$r_h^{\hat{\pi}}(i) = \sum_{a \in \mathcal{A}} \frac{\hat{\lambda}_{h,a}(i)}{\sum_{a' \in \mathcal{A}} \hat{\lambda}_{h,a'}(i)} r_a(i), \qquad P_{h,i}^{\hat{\pi}} = \sum_{a \in \mathcal{A}} \frac{\hat{\lambda}_{h,a}(i)}{\sum_{a' \in \mathcal{A}} \hat{\lambda}_{h,a'}(i)} P_{a,i}.$$

We observe that $r^{\hat{\pi}}$ and $P^{\hat{\pi}}$ are the expected state transition reward and transition probability matrix under the randomized policy $\hat{\pi}$. Note that the value vector $v^{\hat{\pi}}$ under policy $\hat{\pi}$ satisfies
$$r_h^{\hat{\pi}} = v_h^{\hat{\pi}} - P_h^{\hat{\pi}} v_{h+1}^{\hat{\pi}}.$$

It follows that
$$\left(\frac{1}{T}\sum_{k=1}^T \mathcal{G}_k\right) = \sum_{h=0}^{H-1}\sum_{i\in\mathcal{S}}\left(\sum_{a'\in\mathcal{A}}\hat{\lambda}_{h,a'}(i)\right)(r_h^{\hat{\pi}}(i) + P_{h,i}^{\hat{\pi}} v_{h+1}^* - v_h^*(i))$$
$$= \sum_{h=0}^{H-1}\sum_{i\in\mathcal{S}}\left(\sum_{a'\in\mathcal{A}}\hat{\lambda}_{h,a'}(i)\right)\left(e_i(v_h^{\hat{\pi}} - v_h^*) - P_{h,i}^{\hat{\pi}}(v_{h+1}^{\hat{\pi}} - v_{h+1}^*)\right)$$
$$= \left(\sum_{a\in\mathcal{A}}\hat{\lambda}_a\right)^T (\mathbf{I} - \Pi_{\hat{\pi}})(v^{\hat{\pi}} - v^*).$$

Note that $\left(\sum_{a\in\mathcal{A}}\hat{\lambda}_{h,a}(i)\right) \geq \xi_h \geq 1/H^2$ due to the constraint projection step of the SPD-fMDP Algorithm 2. Also note that $r_{h,i}^{\hat{\pi}} + P_{h,i}^{\hat{\pi}} v_{h+1}^* - v_{h,i}^* \geq 0$ for all $i$ and $a$ due to the fact that $v^*$ is the value corresponding to the optimal policy. Then we have
$$\sum_{h=0}^{H-1}\sum_{a\in\mathcal{A}}(\hat{\lambda}_{h,a})^T(r_a + P_a v_{h+1}^* - v_h^*) \geq \frac{1}{H^2}e^T(\mathbf{I} - \Pi_{\hat{\pi}})(v^{\hat{\pi}} - v^*),$$

and each entry of $(\mathbf{I} - \Pi_{\hat{\pi}})(v^{\hat{\pi}} - v^*)$ is greater than 0. By Lemma 7, we have
$$e^T(\mathbf{I} - \Pi_{\hat{\pi}})(v^{\hat{\pi}} - v^*) = \left\|(\mathbf{I} - \Pi_{\hat{\pi}})(v^{\hat{\pi}} - v^*)\right\|_1 \geq \|v^{\hat{\pi}} - v^*\|_\infty.$$

For any $\epsilon_1 > 0, \delta \in (0,1)$, we apply Theorem 7 with $\epsilon = \epsilon_1/H^2$. It gives us
$$\|v^{\hat{\pi}} - v^*\|_\infty \leq H^2\left(\frac{1}{T}\sum_{k=1}^T \mathcal{G}_k\right) = \epsilon,$$

with probability at least $1 - \delta$ where the number of iterations $T$ satisfies
$$T = \Omega\left(\frac{|\mathcal{S}|^4|\mathcal{A}|^2\sigma^2 H^6}{\epsilon^2}\ln\left(\frac{1}{\delta}\right)\right). \tag{27}$$

**Proof of Theorem 8** The proof is similar to the proof of Theorem 5. Denote that $d_{h,a}(i) = r_a(i) + P_{a,i} v_{h+1}^* - v_h^*(i)$. By using the complementarity condition, we have $d_{h,a}(i) > 0$ if $a \neq \pi_h^*(i)$ and $d_{h,a}(i) = 0$ if $a = \pi_h^*(i)$.

We have
$$\mathbf{P}(\hat{\pi}^{Tr} \neq \pi^*) = \mathbf{P}\left(\exists(h,a,i) \text{ s.t. } \pi_h^*(i) \neq a, \hat{\lambda}_{h,a}(i) > \hat{\lambda}_{h,a'}(i), \forall a' \in \mathcal{A}\right)$$
$$\leq \mathbf{P}\left(\sum_{h=0}^{H-1}\sum_{(a,i)}\hat{\lambda}_{h,a}(i)d_{h,a}(i) \geq \frac{(\min_h \xi_h(i))\bar{d}}{|\mathcal{A}|}\right).$$

Let $\epsilon = \frac{(\min_{i \in \mathcal{S}} \xi_i)\bar{d}}{|\mathcal{A}|}$ and apply Theorem 6. We have that for all $\delta > 0$, if

$$T = \Omega\left(\frac{|\mathcal{S}|^4 |\mathcal{A}|^4 H^6 \sigma^2}{\bar{d}^2} \ln\left(\frac{1}{\delta}\right)\right).$$

then

$$\mathbf{P}(\hat{\pi}^{Tr} \neq \pi^*) \leq \mathbf{P}\left(\sum_{h=0}^{H-1} \sum_{(a,i)} \hat{\lambda}_{h,a}(i) d_{h,a}(i) \geq \frac{(\min_h \xi_h(i))\bar{d}}{|\mathcal{A}|}\right) \leq \delta, \quad (28)$$

which completes our proof. ∎